\documentclass{article}
\usepackage{rotating}

\usepackage{ragged2e}       
\usepackage[protrusion=true,expansion=false]{microtype} 
\usepackage{graphicx}
\usepackage{multirow, multicol}
\usepackage{caption}
\usepackage{float}
\usepackage{booktabs, tablefootnote}
\usepackage{tabularx}
\usepackage[table,xcdraw]{xcolor}
\usepackage{amsthm, amsmath,amssymb,amsfonts}
\usepackage{mathtools} 
\usepackage{mathrsfs}
\usepackage{algorithm}
\usepackage{algpseudocode}  

\usepackage{textcomp}
\usepackage{pifont}
\usepackage{setspace,lineno}
\usepackage{listings}
\usepackage{soul}            
\sethlcolor{yellow}          
\usepackage[title]{appendix}
\usepackage[square,numbers]{natbib}
\usepackage{manyfoot}
\usepackage{xcolor}

\definecolor{customgreen}{HTML}{00B050}
\definecolor{captionblue}{HTML}{0070C0}  

\DeclareCaptionLabelFormat{boldblue}{\textbf{\textcolor{captionblue}{#1 #2}}}
\usepackage{titlesec}
\titleformat{\section}
  {\normalfont\Large\bfseries\color{captionblue}}{\thesection}{1em}{}
\titleformat{\subsection}
  {\normalfont\large\bfseries\color{captionblue}}{\thesubsection}{1em}{}
\titleformat{\subsubsection}
  {\normalfont\normalsize\bfseries\color{captionblue}}{\thesubsubsection}{1em}{}

\usepackage{stfloats}

\usepackage{geometry}
\usepackage{amsmath} 
\allowdisplaybreaks[4]
\usepackage[protrusion=true,expansion=false]{microtype}
\usepackage{hyperref}
\hypersetup{
    colorlinks=true,
    linkcolor=blue,
    citecolor=blue,
    urlcolor=blue
}

\usepackage{geometry}
\usepackage{natbib}
\title{ReFRM3D: A Radiomics-enhanced Fused Residual Multiparametric 3D Network with Multi-Scale Feature Fusion for Glioma Characterization}
\author{
Md. Abdur Rahman\textsuperscript{1,2}, 
Md Noman Hossain\textsuperscript{1,2},
Arefin Ittesafun Abian\textsuperscript{1,2}, \\
Mohaimenul Azam Khan Raiaan\textsuperscript{3,*}, 
Yan Zhang\textsuperscript{4}, 
Mirjam Jonkman\textsuperscript{4}, 
Sami Azam\textsuperscript{4,*}\\
\small
\textsuperscript{1}Applied Artificial Intelligence and Intelligent Systems (AAIINS) Laboratory, Dhaka 1217, Bangladesh\\
\small
\textsuperscript{2}Department of Computer Science and Engineering, United International University, Dhaka, 1212, Bangladesh \\
\small
\textsuperscript{3}Department of Data Science and Artificial Intelligence, Monash University, Melbourne, 3800, Australia \\
\small
\textsuperscript{4}Faculty of Science and Technology, Charles Darwin University, Darwin, 0909, Australia \\
\small 
\textsuperscript{*}Corresponding Author: mohaimenul.raiaan@monash.edu; sami.azam@cdu.edu.au
}
\date{} 
\begin{document}
\justifying
\twocolumn[
\maketitle
\begin{abstract}
\noindent 

Gliomas are among the most aggressive cancers, with complex diagnostic processes. Existing glioma segmentation methods often struggle with high variability in imaging data and inadequate optimization. Furthermore, radiomic analysis is typically applied only after segmentation is finished, limiting its ability to inform the segmentation process itself. To address these challenges, we propose a novel radiomics-enhanced fused residual multiparametric 3D network (ReFRM3D) for brain tumor characterization. The framework is based on a 3D U-Net architecture and features multi-scale feature fusion, hybrid upsampling, and an extended residual skip mechanism. Additionally, we introduce a radiomic conditioning mechanism that extracts texture and intensity descriptors from an intermediate coarse segmentation and re-injects them into the decoder to refine the final output. Experimental results on BraTS2019, BraTS2020, and BraTS2021 show strong performance, with mean DSC values of 93.45\%, 93.61\%, and 92.06\%, respectively, across whole tumor, enhancing tumor, and tumor core regions. Compared with recent models, ReFRM3D improves average DSC by 5.79\%, 7.25\%, and 0.96\% on these datasets. Our model also generalized well on the BraTS-Africa dataset with an average DSC of 87.8\%, despite differences in scanner field strength and patient demographics.

\end{abstract}

\vspace{0.5em}
\vspace{1em}
]

\section{Introduction}
\label{sec_introduction}
According to the Central Brain Tumor Registry of the United States, gliomas account for approximately 25\% of all primary brain tumors and represent 80\% of malignant primary tumors of the brain and central nervous system \cite{lei2025fe}. They are classified into distinct grades based on histopathological and molecular characteristics, with low-grade gliomas (LGG) and high-grade gliomas (HGG) being the most commonly investigated types \cite{al2024multi}. Common sequences, such as T1-weighted (T1), T1-weighted contrast-enhanced (T1ce), T2-weighted (T2), and fluid-attenuated inversion recovery (FLAIR), highlight distinct tissue characteristics for tumor progression \cite{lei2025fe}. However, manual segmentation is time-consuming and labor-intensive. Reliable automated segmentation can facilitate lesion localization and morphological quantification, while providing objective information to support therapeutic monitoring \cite{wang2025smart}.

\begin{figure}[h]
\centering
\includegraphics[scale=0.105]{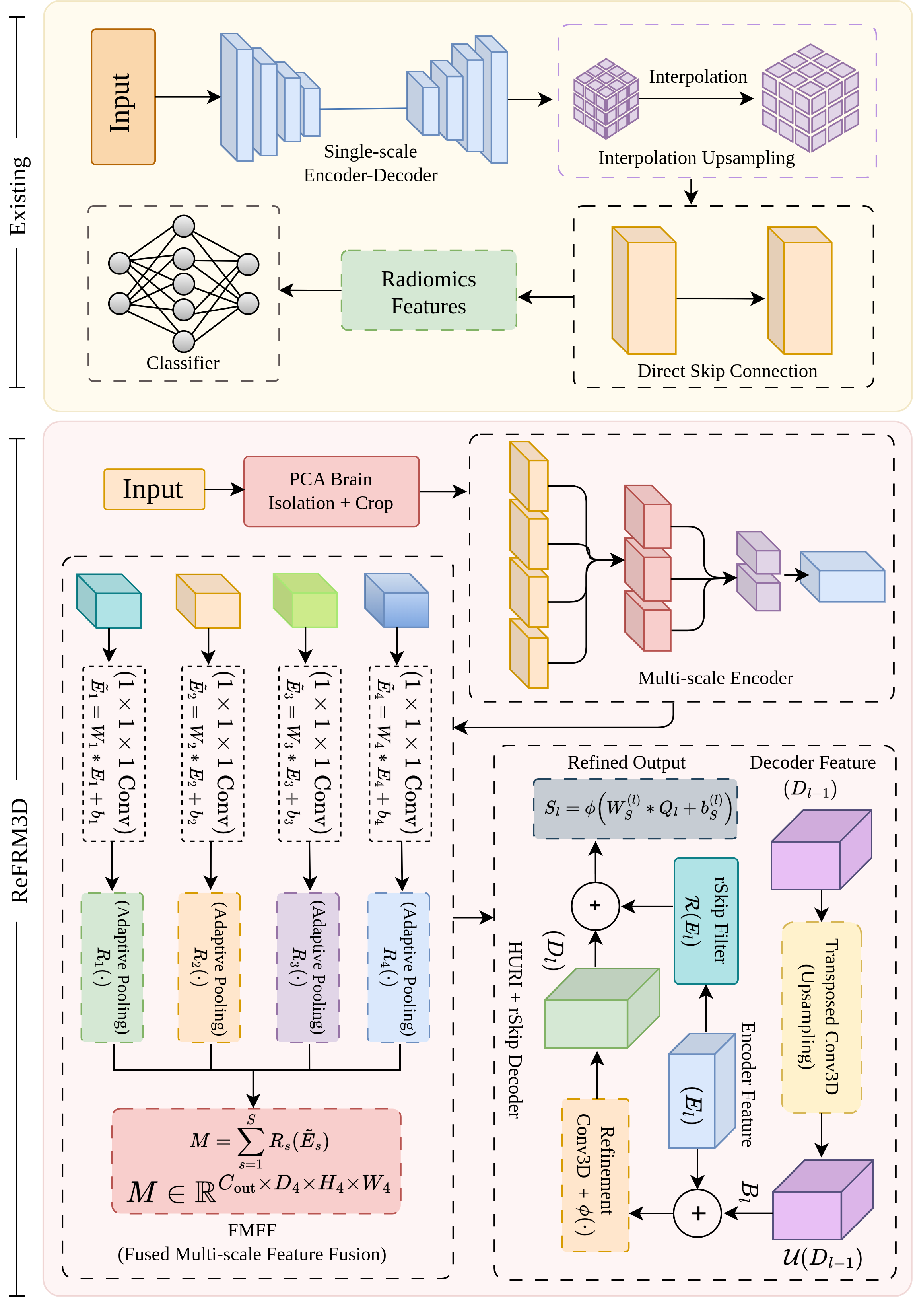}
\caption{Architectural comparison between existing frameworks and the proposed architecture. \textbf{Top}: Standard pipelines utilizing single-scale encoder-decoders, direct skip connections, and basic interpolation. \textbf{Bottom}: The proposed ReFRM3D framework incorporating PCA brain isolation, multi-scale encoding, extended residual skip connections, and a radiomic conditioning loop that extracts descriptors from a coarse mask to guide the final segmentation.}
\label{fig:intro}
\vspace{-7pt}
\end{figure}

{In brain tumor segmentation, 2D deep learning models are limited by their slice-by-slice processing design \cite{zhang2020inter}. By processing each MRI slice independently, 2D convolutional neural networks (CNNs) fail to capture inter-slice spatial dependencies \cite{zhang2020inter}. This structural constraint often leads to inconsistencies in slice-to-slice prediction, discontinuous tumor margins, and irregular boundary distortions after volumetric reconstruction \cite{ali2025drbd}. In contrast, 3D architectures operate directly on voxel-level representations through volumetric kernels. They enable simultaneous modeling of spatial relationships across three orthogonal axes \cite{alsmadi20253d}. The volumetric analysis preserves anatomical continuity, boundary smoothness, and improves volumetric overlap with reference annotations \cite{saleh2025traditional}. Despite substantial computational and memory requirements, 3D approaches demonstrate superior geometric accuracy and more consistent subregion definition for clinical brain tumor segmentation.}

Numerous image processing and machine learning approaches \cite{kumar2026comprehensive, li2024multi, zhang2021msmanet, sultan2026deep} have been developed to enhance brain tumor segmentation. Among these methodologies, feature fusion techniques, attention mechanisms, and multi-modal data integration are widely applied for dense connectivity \cite{zhang20213d, zhou2021erv, ding2020multi, li2019novel}, probabilistic graph-based voxel label fusion, window-modulated attention, and hypergraph membrane \cite{zhong20202wm, xue2020hypergraph} systems to support feature extraction and precise boundary identification. Customized loss functions and hybrid optimization strategies are also used in multi-scale and hierarchical training techniques to optimize segmentation learning \cite{barzegar2021wlfs, zhong20202wm, liu2023multiscale}. To maximize computational efficiency, model architectures employ lightweight network designs, optimized convolutional operations, and efficient encoder structures \cite{xu2022brain}.

Despite these advancements, many frameworks still encounter several limitations. Current pipelines predominantly rely on single-scale feature representations and standard interpolation upsampling (see Figure \ref{fig:intro}), which often fail to preserve fine-grained topological boundaries in complex 3D tumor structures.

To overcome these challenges, we propose the ReFRM3D framework, an integrated 3D architecture engineered to systematically bridge these gaps. The major contributions of our study are as follows:

\begin{itemize}
\setlength{\itemsep}{0pt}
    \item A novel approach for isolating and cropping brain regions from 3D scans using intensity thresholding and connected component analysis is introduced to minimize computational load while maintaining brain structures for tumor localization.
    \item A voxel intensity standardization technique is developed to reduce imaging variability and to normalize data for multi-grade tumors (i.e., high-grade and low-grade glioma).
    \item To enhance multi-scale feature fusion, improve localization, and reduce interpolation artifacts in tumor segmentation, an extended 3D U-Net is proposed with augmented Fused Multi-scale Feature Fusion, Hybrid Upsampling, Residual Integration, and the Residual Skip Mechanism.
    \item A radiomic conditioning mechanism is introduced that extracts texture and intensity descriptors from an intermediate coarse segmentation, and re-injects them into the decoder to guide the final output.
\end{itemize}

The rest of the paper is organized as follows: Section \ref{sec_related_work} reviews related work. Section \ref{sec_methodology} describes the methodology, including preprocessing, segmentation, and radiomic conditioning. Section \ref{sec_analysis_of_results} presents results analysis and comparisons. Section \ref{sec_discussion} provides a discussion of our study, followed by the conclusion in Section \ref{conclusion}.

\section{Related Works}
\label{sec_related_work}

\noindent\textbf{Segmentation Architectures.} Existing studies explored feature representation through dense connectivity, multi-scale feature fusion, and adaptive feature aggregation to complement low- and high-level features \cite{zhang20213d, zhou20203d, zhou2021erv, ding2020multi, li2019novel}. For example, Zhang et al. \cite{zhang20213d} integrated multi-view 2D-CNN predictions with fused modalities to improve spatial feature representation, while Zhou et al. \cite{zhou20203d, zhou2021erv} employed dense connectivity and atrous convolutions to enhance multi-scale contextual learning. Similarly, Ding et al. \cite{ding2020multi} introduced adaptive fusion blocks to preserve complementary low- and high-level information, whereas Li et al. \cite{li2019novel} incorporated inception modules to capture diverse features across multiple receptive fields for improved local and global tumor representation.

Beyond general feature fusion, several studies addressed tumor boundary delineation. For instance, Tong et al. \cite{tong2023dual} separated global and local feature representations to improve tumor structure modeling, while Jiang et al. \cite{jiang2021novel} employed dual-stream decoders to enhance edge-aware feature learning and delineate tumor boundaries more accurately. Attention-based approaches have further been explored to improve segmentation in ambiguous or difficult tumor regions. For example, Liu et al. \cite{liu2023attention} incorporated attention mechanisms to refine feature selection in challenging areas. Similarly, Ullah et al. \cite{ullah2022cascade} adopted progressive cascade refinement to iteratively improve segmentation accuracy, while Liu et al. \cite{liu2023multiscale} combined multi-scale attention with hierarchical decoupled convolutions to capture both complex and fine tumor characteristics. Zhong et al. \cite{zhong20202wm} also introduced window-modulated attention to emphasize difficult tumor regions that are difficult to distinguish with conventional convolutions.

Despite these architectural advances, most existing approaches primarily focus on improving segmentation accuracy through increasingly sophisticated feature extraction and fusion mechanisms. Moreover, some methods operate directly on complete MRI volumes without explicitly isolating the brain region beforehand. Such full-volume processing may introduce computational redundancy by allocating model capacity to non-informative background regions. Importantly, segmentation is generally treated as the primary endpoint, with limited consideration of how segmentation-derived representations can subsequently support quantitative tumor characterization and subtype classification.

\paragraph{Multimodal Data Fusion.} Multimodal MRI fusion has been widely adopted in glioma segmentation as different MRI sequences reveal complementary information about tumor appearance, tissue composition, and anatomy. Existing approaches have explored fusion at different stages, including voxel-level label integration, feature-level fusion, ensemble learning, and attention-based multi-slice representation. 

For example, Barzegar et al. \cite{barzegar2021wlfs} employed probabilistic graph-based voxel-label fusion to combine prior atlas information with target-image characteristics for improved voxel-wise label assignment. In Addition, Xue et al. \cite{xue2020hypergraph} introduced a hypergraph membrane framework for multimodal ensemble learning to model complex relationships among tumor structures across MRI sequences. Xu et al. \cite{xu2022brain} also incorporated corner attention to capture spatial relationships across multiple slices and improve the representation of complementary multimodal information.

Other studies have combined multimodal information with cascaded architectures, attention mechanisms, and preprocessing strategies. Ranjbarzadeh et al. \cite{ranjbarzadeh2021brain} employed cascade CNNs with distance-wise attention to progressively refine tumor boundaries, while Li et al. \cite{li2022automatic} incorporated preprocessing strategies to improve input representation and, consequently, segmentation performance. The nnU-Net framework proposed by Isensee et al. \cite{isensee2021nnu} additionally demonstrated the importance of dataset-specific configuration, post-processing, and ensemble strategies across heterogeneous glioma datasets. In another study, Li et al. \cite{li2024multi} explored multimodal feature learning to improve the representation of complementary information across MRI sequences.

Although these studies demonstrate the value of multimodal information, the primary objective of most fusion strategies remains segmentation improvement. In particular, voxel-level, feature-level, and ensemble-based fusion approaches \cite{barzegar2021wlfs, xue2020hypergraph} primarily seek more discriminative tumor representations for voxel-wise prediction rather than explicitly exploiting those representations for tumor subtype classification. Likewise, preprocessing- and attention-based approaches \cite{ranjbarzadeh2021brain, li2022automatic} improve segmentation quality but do not establish a unified pathway from multimodal segmentation to downstream radiomic analysis.

\begin{figure*}[h]
\centering
\includegraphics[scale=0.078]{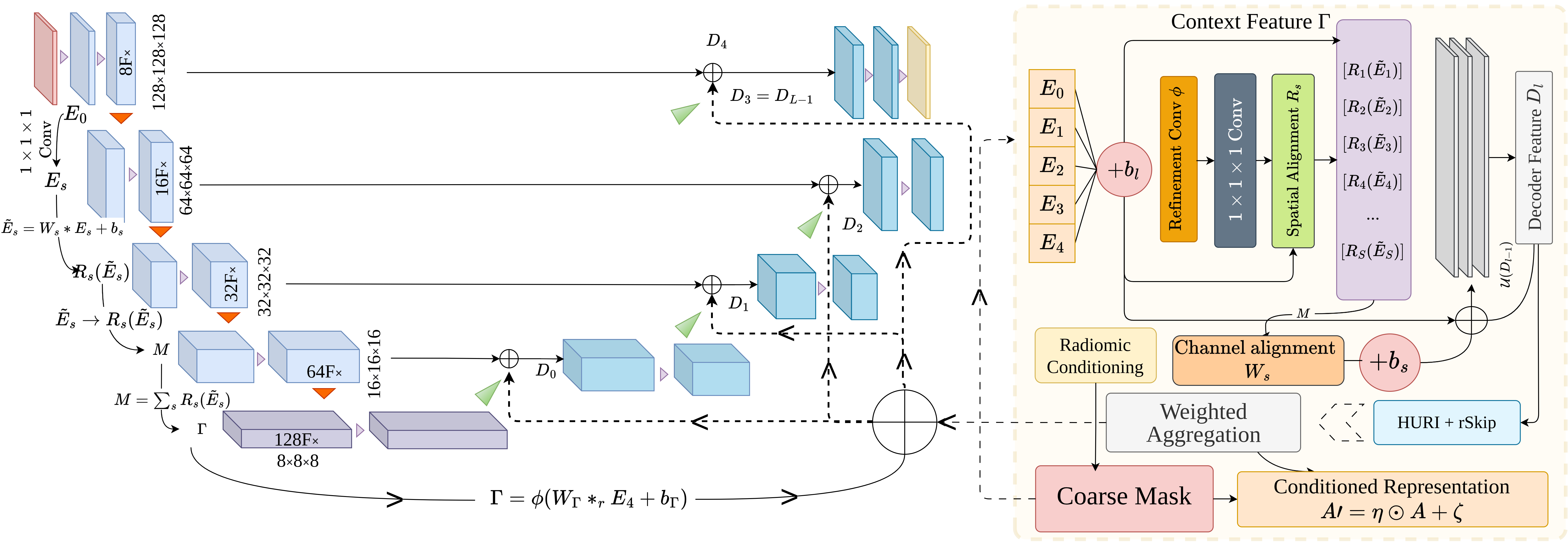}
\caption{Overview of the proposed glioma segmentation framework. The expansion layers mirror the contraction layers in size and filters.}
\label{fig:refrm3d}
\end{figure*}

\paragraph{Radiomics-based Tumor Analysis.} Radiomics provides a complementary approach to deep feature learning by enabling the high-throughput extraction of quantitative descriptors related to tumor shape, intensity, and texture. Van Griethuysen et al. \cite{van2017computational} established a standardized computational framework for radiomic feature extraction and demonstrated its applicability to a range of brain tumor tasks, including tumor grading, prognosis, and treatment-response assessment. 

Recent studies have further demonstrated the potential of radiomic features for glioma characterization and clinical applications \cite{wang2025clinical}. Sun et al. \cite{sun2024glioma} combined automatic tumor segmentation using nnU-Net with subsequent radiomic feature extraction to predict adult-type diffuse glioma subtypes. Li et al. \cite{li2023universal} proposed TFCS, a feature-selection strategy that jointly considers correlation, redundancy, and model-specific roles to identify informative radiomic feature subsets for brain tumor grade prediction. Similarly, Ali et al. \cite{ali2026rgnn3d} integrated PyRadiomics-derived features with graph convolutional and Long Short-Term Memory layers to construct patient-level graphs from 3D MRI radiomic representations for low- and high-grade gliomas. For pediatric gliomas, Liu et al. \cite{liu2024early} investigated the prognostic value of radiomic features extracted from segmented tumor volumes by training Support Vector Machine classifiers to predict overall survival.

More recent hybrid approaches have attempted to combine handcrafted radiomic descriptors with learned deep representations. Taleb et al. \cite{taleb2026hybrid} integrated deep features obtained from EfficientNet-B0 and ResNet-18 with PyRadiomics features using sequential feature selection before classification. Similarly, Rastogi et al. \cite{rastogi2025deep} combined a 3D replicator autoencoder with handcrafted morphological features extracted from segmented tumor regions for survival-oriented representation learning.

Despite these developments, radiomic analysis is commonly implemented as a downstream stage, in which features are extracted only after tumor segmentation is complete, and subsequently supplied to an independent classification or prognostic model. This dependence on a finished segmentation means radiomic descriptors cannot inform the segmentation process itself.

\section{Methodology}
\label{sec_methodology}
Figure \ref{fig:refrm3d} illustrates the overall proposed segmentation framework. The proposed ReFRM3D model combines several key components to address the challenges in glioma segmentation.

\subsection{Datasets}
\label{subsec_datasets}
We used four benchmark datasets from the Brain Tumor Segmentation (BraTS) challenge series \cite{menze2014multimodal, baid2021rsna}.

\vspace{5pt}
\noindent\textbf{BraTS 2019--2021 datasets.} BraTS2019 comprises 3D multiparametric MRI (mpMRI) scans in NIfTI format from 626 glioma patients, including high-grade glioma (HGG) and lower-grade glioma (LGG), with 355 publicly available samples containing ground-truth (GT) masks. BraTS2020 extends this collection to 660 cases, with 369 publicly available samples and GT masks. BraTS2021 further expands the dataset to 2,040 patients, including 1,251 samples with GT masks. All three versions contain four MRI sequences: T1-weighted (T1), contrast-enhanced T1-weighted (T1ce/T1gd), T2-weighted (T2), and T2-FLAIR (FLAIR).

\vspace{5pt}
\noindent{\textbf{BraTS-Africa dataset.} To evaluate generalizability, we additionally validated our model on the BraTS-Africa dataset \cite{adewole2025brats}. The dataset consists of 146 cases collected from six diagnostic centers in Nigeria between January 2010 and December 2022. It includes 95 patients with adult diffuse glioma and 51 patients with other central nervous system neoplasms, acquired on 1.5T MRI.}

Since T1gd/T1ce is the contrast-enhanced form of T1, we use T1gd/T1ce together with T2 and FLAIR, and exclude the native T1 sequence.

\subsection{Preprocessing}
\label{subsec_preprocessing}
The images were resized to $128\times128\times128$ with three channels (i.e., FLAIR, T1ce/T1gd, T2). {The ground truth segmentation masks are used only during training for slice selection optimization. During validation, the complete MRI volumes are processed without requiring ground truth masks.} The following sections detail the preprocessing steps for optimizing MRI data, including brain region isolation, voxel intensity standardization, and slice range selection. The preprocessing pipeline is outlined in Algorithm \ref{algo:glioma_preprocessing_algo}.

\begin{algorithm}[h]
\caption{Data Preprocessing Pipeline}
\label{algo:glioma_preprocessing_algo}
\textbf{Input:} 3D Volumes $V = \{v_1, \ldots, v_n\}$, Segmentation Masks $S = \{s_1, \ldots, s_n\}$ \\
\textbf{Key Parameters:} Intensity Threshold $T_i$, Margin $\delta$, Covariance Matrix $\mathbf{C}$, Principal Axes $Q$, Mean $\mu$, Std. Dev. $\sigma$, Z-score $Z$

\begin{algorithmic}[1]
    \State \textit{\color{gray}// Brain region isolation and cropping}
    \ForAll{$v \in V$}
        \State $T_i \gets \text{percentile}(v, 99)$
        \State $M \gets \text{binary\_mask}(v, T_i)$
        \State $X \gets \text{largest\_component}(\text{connected\_comp.}(M))$
        \State $\mathbf{C} \gets \text{covariance\_matrix}(X)$
        \State $Q \gets \text{eigenvectors}(\mathbf{C})$
        \State $v_{\text{cropped}} \gets \text{crop\_pca}(v, Q, 1.5\sigma)$
    \EndFor
    
    \State $V_{\text{cropped}} \gets \{v_{\text{cropped}_i} \mid i = 1, \ldots, n\}$

    \State \textit{\color{gray}// Intensity standardization}
    \ForAll{$v_{\text{cropped}} \in V_{\text{cropped}}$}
        \State $\mu, \sigma \gets \text{mean\_std}(v_{\text{cropped}})$ 
        \State $Z \gets \text{zscore\_normalize}(v_{\text{cropped}}, \mu, \sigma)$ 
    \EndFor

    \State $V_{\text{norm}} \gets \{Z_i \mid i = 1, \ldots, n\}$

    \State \textit{\color{gray}// Slice range selection}
    \ForAll{$s \in S$}
        \State $S_{\text{start}} \gets \min \{z \mid \max(s_z) > 0\}$
        \State $S_{\text{end}} \gets \max \{z \mid \max(s_z) > 0\}$
        \State $v_{\text{sliced}} \gets \text{slice\_volume}(v_{\text{norm}}, S_{\text{start}}, S_{\text{end}})$
    \EndFor

    \State $V_{\text{sliced}} \gets \{v_{\text{sliced}_i} \mid i = 1, \ldots, n\}$

    \State \textit{\color{gray}// Resize to fixed dimensions}
    \ForAll{$v_{\text{sliced}} \in V_{\text{sliced}}$}
        \State ${\text{Vol}}[i] \gets \text{resize}(v_{\text{sliced}}, 128^3)$ 
        \State ${\text{Vol}}_{\text{data}}[i] \gets \text{to\_npy}(\text{Vol}[i])$
    \EndFor
\end{algorithmic}
\textbf{Output:} Preprocessed Data, ${\text{Vol}}_{\text{data}}[i]$
\end{algorithm}

\subsubsection{Brain Region Isolation}
Traditional methods select a fixed rectangular region and risk removing brain tissue or including non-brain regions. We designed an isolation method to automatically account for the multi-planar nature of MRI data across axial, coronal, and sagittal views.

We first normalize intensity values across scans for consistency. We then generate an initial binary mask \( M \) using intensity thresholding, with the threshold \( T \) set at the 99th percentile of the voxel intensity distribution \( V \). The 99th percentile separates the highest 1\% of voxel intensities from the rest, and we compute it independently for each scan after normalization. Equation \eqref{eq:voxel_intensity} defines this step:
\begin{equation}
    M(i, j, k) = 
    \begin{cases} 
      1 & \text{if } V(i, j, k) > T \\
      0 & \text{otherwise}
   \end{cases}
   \label{eq:voxel_intensity}
\end{equation}
where \( M(i, j, k) \) indicates the presence (1) or absence (0) of brain tissue at voxel \( (i, j, k) \). We treat this mask as an initial approximation of the brain region. Next, we apply connected component analysis and retain only the largest connected component to remove high-intensity artifacts. In the BraTS datasets, the brain forms a single contiguous structure within the surrounding background. Retaining the largest component therefore isolates the brain while preserving tumors, since tumors remain part of this connected structure regardless of location or intensity.

We then extract the voxel coordinates of this region as \( X \), and apply Principal Component Analysis (PCA) to define the brain's bounding box. Axis-aligned cropping assumes a fixed brain orientation across scans, but brain orientation varies across patients and acquisitions. PCA instead adapts the bounding box to the actual orientation of the brain, computed directly from the voxel distribution. We compute the covariance matrix \( \mathbf{C} \) as:
\begin{equation}
    \mathbf{C} = \frac{1}{n} \sum_{i=1}^{n} (X_i - \bar{X})(X_i - \bar{X})^T
    \label{eq:covariance_matrix}
\end{equation}
where \( \bar{X} \) is the mean of the voxel coordinates. The eigenvectors \( Q_j \) of \( \mathbf{C} \) give the principal axes of the brain. We then project each voxel coordinate onto these eigenvectors to find the minimum and maximum extents along each axis:
\begin{equation}
    \begin{aligned}
        x_{\text{min}} &= \min_{i=1,\dots,n}(X_i \cdot Q_x), \quad x_{\text{max}} = \max_{i=1,\dots,n}(X_i \cdot Q_x), \\
        y_{\text{min}} &= \min_{i=1,\dots,n}(X_i \cdot Q_y), \quad y_{\text{max}} = \max_{i=1,\dots,n}(X_i \cdot Q_y), \\
        z_{\text{min}} &= \min_{i=1,\dots,n}(X_i \cdot Q_z), \quad z_{\text{max}} = \max_{i=1,\dots,n}(X_i \cdot Q_z)
    \end{aligned}
    \label{eq:pca_bounds}
\end{equation}
where \( Q_x \), \( Q_y \), and \( Q_z \) are the eigenvectors of \( \mathbf{C} \) along each axis, and \( X_i \cdot Q_x \) denotes the scalar projection of \( X_i \) onto that axis. Each eigenvector gives a principal direction that aligns with the brain's shape.

Finally, we add a margin of 1.5 times the standard deviation ($1.5\sigma$) of the voxel coordinates along each principal axis to account for anatomical variability. This margin-adjusted bounding box crops the brain region accurately, without including irrelevant non-brain regions (see Section \ref{preproc_slice}).

\subsubsection{Intensity Standardization and Slice Range}
\label{subsec_intensity_slice}
After brain region isolation and cropping, we apply Z-score normalization to standardize voxel intensity values across MRI volumes. This step corrects intensity variation across acquisition sequences, namely T1ce/gd, T2, and FLAIR. We use the brain region mask and restrict the intensity calculation to this relevant anatomical region.

We first extract voxel intensities within the brain region and compute the mean and standard deviation as shown in Equations \eqref{mean_eq} and \eqref{std_dev_eq}:
\begin{equation}
\mu = \frac{1}{N} \sum_{i=1}^{N} v_i
\label{mean_eq}
\end{equation}
\begin{equation}
\sigma = \sqrt{\frac{1}{N} \sum_{i=1}^{N} (v_i - \mu)^2}
\label{std_dev_eq}
\end{equation}
where \(v_i\) is the intensity of voxel $i$, \(\mu\) and \(\sigma\) are the mean and standard deviation of these intensities, and \(N\) is the total number of voxels in the brain region.

\begin{figure}[!ht]
    \centering
    \includegraphics[scale=0.06]{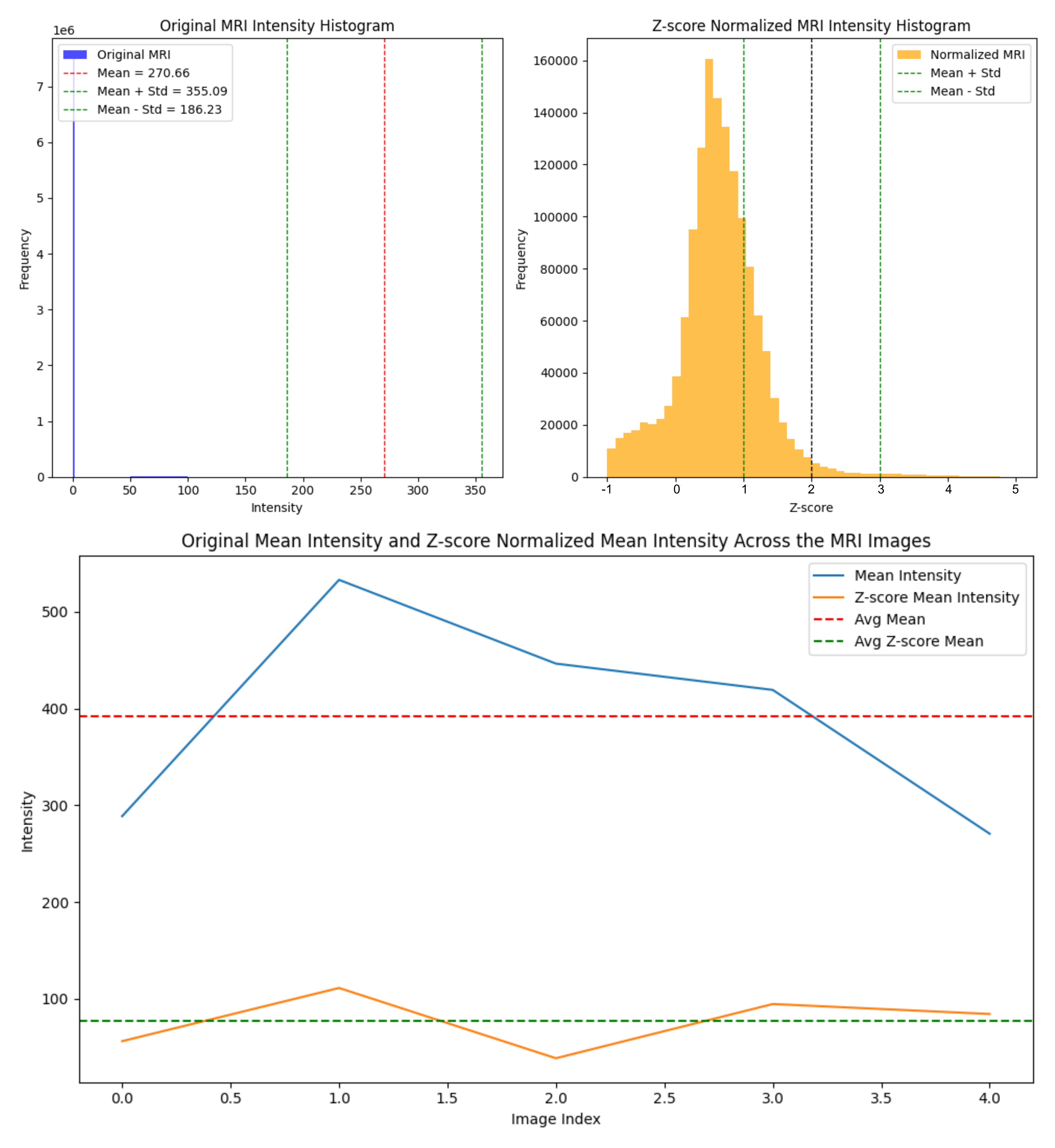}
    \caption{Visualization of histograms of original and Z-score normalized intensity distributions with intensity comparison across multiple MRI scans. {The top-left panel shows the original intensity histogram with high variability (mean \(\mu = 270.66\), standard deviation \(\sigma = 84.43\)), while the top-right panel displays the normalized distribution centered at zero with unit variance. The bottom panel illustrates that while original mean intensities fluctuate significantly across scans (ranging from 290 to 550), the normalized values remain stable (70 to 110), which also ensures consistent input for model training.}}
    \label{fig:z_score}
    \vspace{-5pt}
\end{figure}

We then apply Z-score normalization to center intensities at zero and scale them to unit standard deviation. For voxel intensity \( v_{ijk} \), we compute the normalized value \( Z_{ijk} \) as in Equation \eqref{eq:z_score_eq}:
\begin{equation}
    Z_{ijk} = \frac{v_{ijk} - \mu}{\sigma}
    \label{eq:z_score_eq}
\end{equation}
Figure \ref{fig:z_score} compares the original and normalized intensity distributions.

We note that certain MRI slices contain no tumor region, and including them adds computational cost without benefit. We therefore isolate tumor-containing slices and restrict training to this reduced range. We check each slice of the segmentation mask \( S \) for non-zero values, and define the start and end of the tumor region as:
\begin{equation}
    S_{\text{start}} = \min\{ s \mid \max(S_s) > 0 \}
\end{equation}
\begin{equation}
    S_{\text{end}} = \max\{ s \mid \max(S_s) > 0 \}
\end{equation}
where \( S_{\text{start}} \) and \( S_{\text{end}} \) are the first and last slice indices containing tumor. We then restrict the depth dimension \( D \) of the volume \( (D, H, W) \) to the range \( [S_{\text{start}}, S_{\text{end}}] \), giving a reduced size of \( (S_{\text{end}} - S_{\text{start}} + 1, H, W) \).

Since BraTS modalities are co-registered, this slice range applies uniformly across all sequences. We identify a tumor-containing slice from any non-zero voxel, so even small lesions are retained. We also keep all slices between the first and last tumor slice to preserve surrounding anatomical context rather than selecting slices aggressively.

\begin{figure}[!ht]
    \centering
    \includegraphics[scale=0.11]{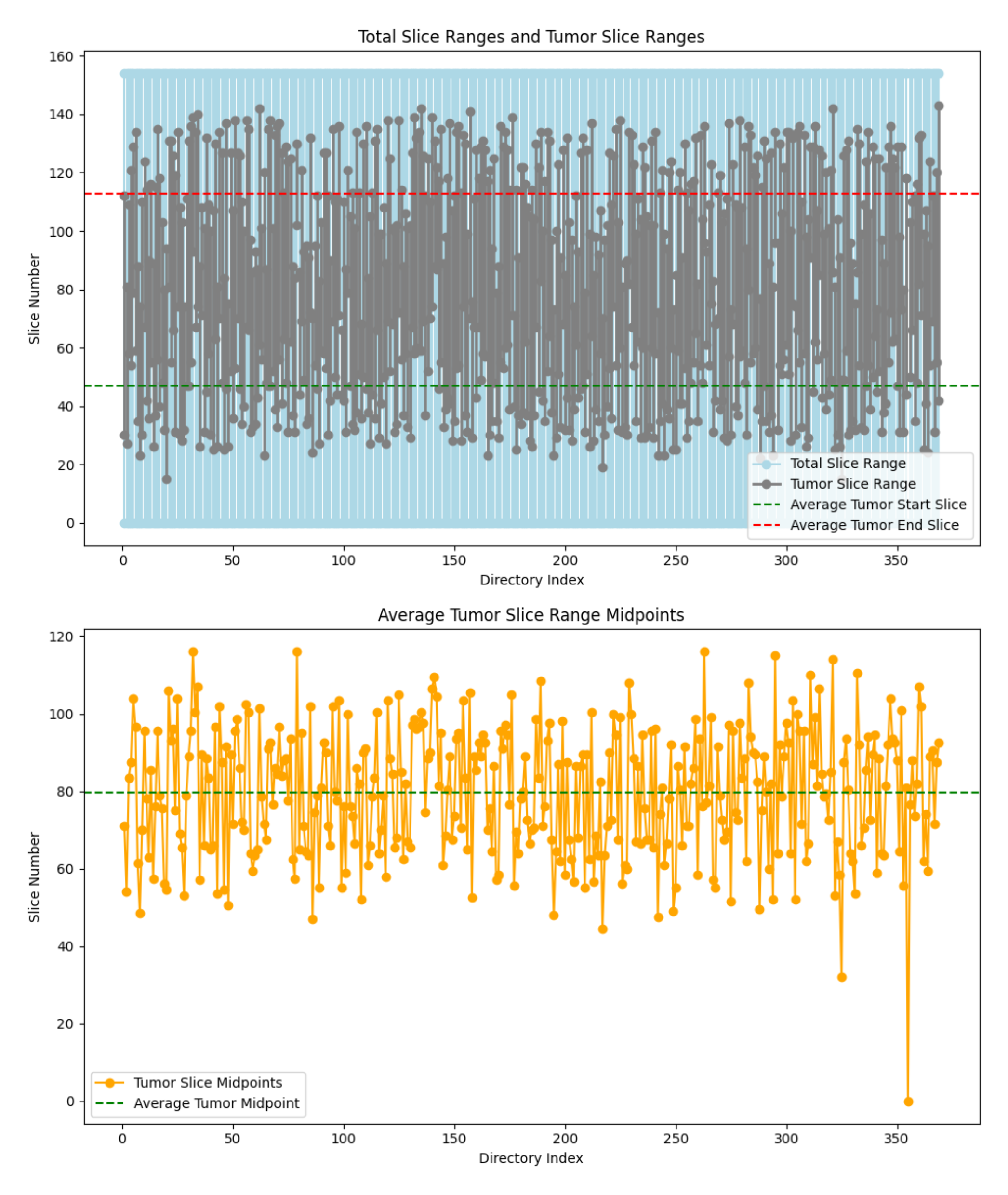}
    \caption{Visualization of total slice ranges. Tumors consistently appear between slices 46 and 113 for all 369 volumes (BraTS2020). On average, they occupy 68 slices per volume. The top panel shows tumor slice ranges (gray lines) within the total volume range (light blue bars). Average start and end positions are marked by green and red dashed lines, respectively. The bottom panel displays tumor midpoint positions and shows anatomical consistency across patients, with an average center at slice 79.5. This analysis confirms that excluding non-tumor slices reduces the processed slice count by approximately 56\% (from 155 to 68 slices) without affecting segmentation performance.}
    \label{fig:slice_range}
    \vspace{-10pt}
\end{figure}

Figure \ref{fig:slice_range} shows the selected slice ranges for this experiment. Restricting training to these slices caused no performance loss and reduced computational cost (see Section \ref{preproc_slice}).

\subsection{Segmentation Model}
\label{subsec_segmentation_mode}
\begin{algorithm}[!ht]
\caption{Radiomic Conditioning-based Segmentation}
\label{algo:glioma_segmentation_algo}
\textbf{Params:} 3D Volumes $V = \{v_1, \ldots, v_n\}$, Segment Mask $\hat{Y}$, Radiomic Embedds $Z$, Coarse Mask $\hat{Y}_{\text{coar}}$, Kernel $k$, Dilation $r$, Encoder Stages $L_S=4$, Fusion Scales $F_S=4$, Learnable Weights $\gamma_l, W_R^{(l)}, b_R^{(l)}, W_p, W_\eta, W_\zeta, b_p, b_\eta, b_\zeta$.
\begin{algorithmic}[1]
    \ForAll{$v \in V$}
        \State \textit{\color{gray}// Encoder: downsampling stages}
        \State $E_0 \gets v$
        \For{each encoder stage $l = 1$ \textbf{to} $L_S$}
            \State $E_l \gets \text{mxpool}(\text{relu}(\text{conv3d}_{k}(E_{l-1})))$
        \EndFor
        
        \State $\Gamma \gets \text{relu}(\text{dilated\_conv3d}_{r}(E_4))$ \textit{\color{gray}// Context pathway}
        
        \State \textit{\color{gray}// FMFF: fuse all encoder stages}
        \For{each scale $s = 1$ \textbf{to} $F_S$}
            \State $\tilde{E}_s \gets \text{conv1x1}(E_s)$
            \State $\tilde{E}_s \gets \text{resize}(\tilde{E}_s, \text{size}(E_4))$
        \EndFor
        \State $M \gets \sum_{s=1}^{F_S} \tilde{E}_s$
        
        \State \textit{\color{gray}// Decoder: upsampling stages w/ HURI}
        \State $D_4 \gets \text{relu}(\text{conv3d}(\text{upsamp}(\Gamma+M) + \text{conv1x1}(E_4)))$
        \For{each decoder stage $l = 3$ \textbf{downto} $1$}
            \State $D_l' \gets \text{conv3d}(\text{upsamp}(D_{l+1}) + \text{conv1x1}(E_l))$
            \State $D_l \gets \text{relu}(D_l')$
        \EndFor
        
        \State \textit{\color{gray}// Coarse mask and radiomics}
        \State $\hat{Y}_{\text{coar}} \gets \text{sigmoid}(\text{conv1x1}(D_1))$
        \State $R \gets \text{extract\_radiomics}(\hat{Y}_{\text{coar}}, v)$
        \State $Z \gets W_p \cdot \text{normalize}(R) + b_p$
        \State $\eta \gets W_\eta Z + b_\eta$
        \State $\zeta \gets W_\zeta Z + b_\zeta$
        
        \State \textit{\color{gray}// rSkip: aggregate all decoder-encoder pairs}
        \State $A \gets \mathbf{0}$
        \For{each skip conn $l = 1$ \textbf{to} $4$}
            \State $S_l \gets D_l + \text{conv1x1}_{W_R^{(l)}}(E_l)$
            \State $A \gets A + \gamma_l \cdot S_l$
        \EndFor
        
        \State \textit{\color{gray}// Apply conditioning}
        \State $A' \gets \eta \odot A + \zeta$
        \State $\hat{Y} \gets \text{softmax}(\text{conv1x1}(A'))$
    \EndFor
    
    \State \textbf{return} $\hat{Y}, Z, \hat{Y}_{\text{coar}}$
\end{algorithmic}
\end{algorithm}

In this section, we detail the proposed segmentation model for glioma segmentation. The overall pipeline is outlined in Algorithm \ref{algo:glioma_segmentation_algo}.


\subsubsection{Base Model and Context Pathway}
\label{subsec_base_context}
Our backbone builds on an encoder-decoder structure. At each encoder layer, we apply convolution, activation, and max pooling with stride $2 \times 2 \times 2$, using kernel size \( k = 3 \) throughout except at the final output layer, where \( k = 1 \). Repeated downsampling expands the receptive field of the encoder. Its size at layer $l$ follows the recurrence:
\begin{equation}
\rho_l = \rho_{l-1} + (k-1)\prod_{j=1}^{l-1} s_j, \qquad \rho_0 = 1
\label{eq:receptive_field_enc}
\end{equation}
where $s_j$ is the stride at layer $j$. The decoder mirrors this structure through upsampling.

Equation \eqref{eq:receptive_field_enc} shows the encoder's receptive field grows through repeated downsampling, which reduces spatial resolution. To gather global context without this cost, we introduce a context pathway that applies a dilated convolution to the final encoder output \( E_L \), with dilation rate $r$ and no further downsampling. This produces the context feature map $\Gamma$, as shown in Equation \eqref{eq:context_pathway}:
\begin{equation}
\Gamma = \phi\big(W_\Gamma *_r E_L + b_\Gamma\big)
\label{eq:context_pathway}
\end{equation}
where $W_\Gamma, b_\Gamma$ are the learnable weights and bias of the context pathway, and $*_r$ denotes convolution sampled at spacing $r$ rather than at unit spacing. Since dilation widens the sampling spacing rather than the kernel footprint, $\Gamma$ retains the spatial resolution of $E_L$ while covering a wider receptive field. Stacked dilated convolutions are known to suffer from the gridding problem, where the kernel skips voxels and leaves gaps in the sampled input. As given in Equation \eqref{eq:gridding_condition}, a dilation rate $r$ avoids this only if it satisfies the coverage condition
\begin{align}
M_i &= \max\left(M_{i+1} - 2r_i,\; M_{i+1} - 2(M_{i+1} - r_i),\; r_i\right) \notag \\
M_n &= r_n, \qquad M_2 \le k
\label{eq:gridding_condition}
\end{align}
Under this coverage condition, the dilation provides an effective kernel size while leaving the parameter count unchanged, as shown in Equation \eqref{eq:eff_kernel_params}:
\begin{equation}
k_{\text{eff}} = r(k-1) + 1, \qquad |\theta_\Gamma| = k^3 \, C_L \, C_{\text{out}}
\label{eq:eff_kernel_params}
\end{equation}
Combining the encoder's pooling-driven growth with the dilation term, as shown in Equation \eqref{eq:receptive_field_total}, gives the total receptive field reaching the final context output:
\begin{equation}
\rho_{\text{total}} = \rho_L + r(k-1)\prod_{j=1}^{L} s_j
\label{eq:receptive_field_total}
\end{equation}
which quantifies how far a single voxel's influence extends across the volume once both encoder depth and dilation are accounted for.

\subsubsection{Fused Residual Backbone}
\label{subsec_fused_residual_backbone}
A plain encoder-decoder pair loses fine detail as it deepens, since single-scale fusion, plain upsampling, and untransformed skip connections each have distinct limitations. We address this through a fused backbone, composed of three components: Fused Multi-scale Feature Fusion (FMFF), Hybrid Upsampling and Residual Integration (HURI), and a Residual Skip connection (rSkip).

\begin{figure*}[!ht]
\centering
\includegraphics[scale=0.08]{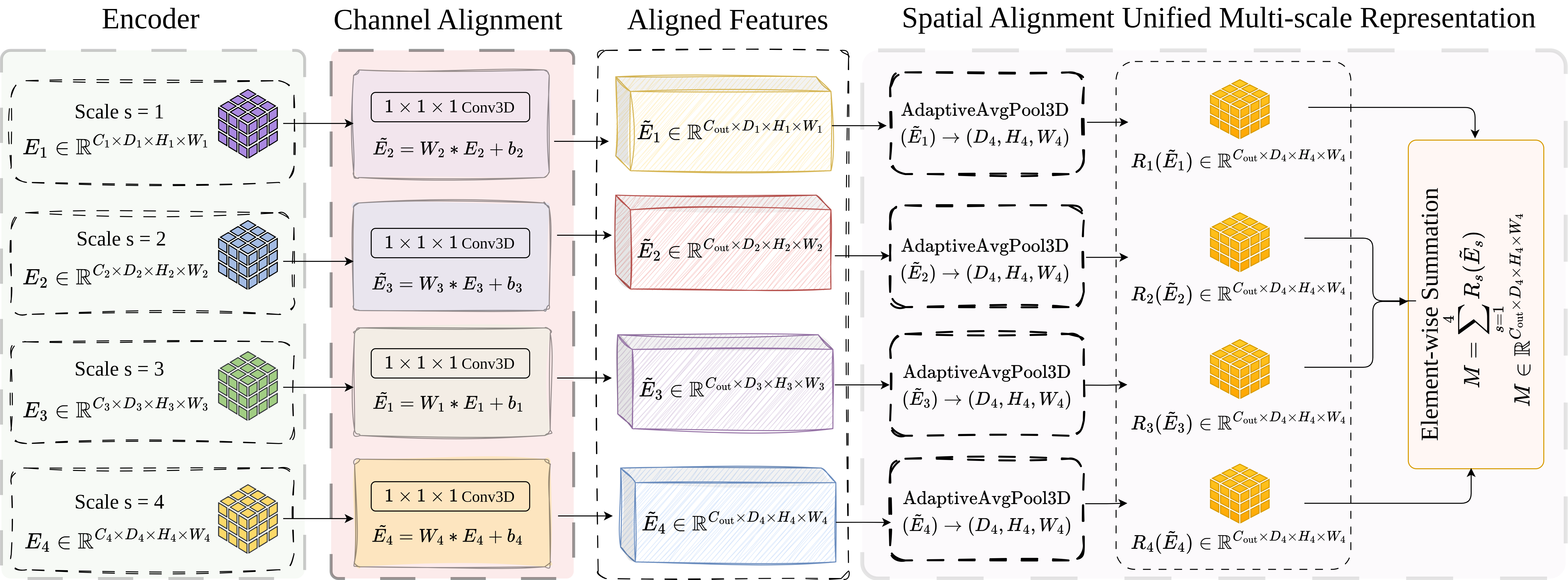}
\caption{Overview of the Fused Multi-scale Feature Fusion (FMFF) module. Encoder features from multiple scales undergo channel alignment via $1\times1\times1$ convolutions and spatial alignment via adaptive pooling to form a unified representation.}
\label{fig:fmff}
\end{figure*}

\paragraph{Fused Multi-scale Feature Fusion.} Encoder stages at different depths differ in both channel count and spatial resolution. In our approach (see Figure \ref{fig:fmff}), we first align each encoder stage $s \in \{1,\ldots,S\}$ to a common channel dimension through a learnable $1\times1$ convolution,
\begin{equation}
\tilde{E}_s = W_s * E_s + b_s
\label{eq:fmff_align}
\end{equation}
We then resize each aligned map to a shared spatial resolution through the operator $R_s(\cdot)$, since deeper stages are spatially smaller than shallow ones. The fused representation follows Equation \ref{eq:fmff_fuse}:
\begin{equation}
M = \sum_{s=1}^{S} R_s(\tilde{E}_s)
\label{eq:fmff_fuse}
\end{equation}
Since each term is aligned in channel and resolution before summation, the fused map inherits the largest receptive field among all contributing scales, rather than an average or a sum of them, as shown in Equation \ref{eq:fmff_rf}:
\begin{equation}
\rho_M = \max_{s=1,\ldots,S} \rho_s
\label{eq:fmff_rf}
\end{equation}
As a result, FMFF broadens contextual coverage while preserving fine-scale detail. The fused map $M$ combines with the context feature $\Gamma$ at the bottleneck, so the decoder's first upsampling stage draws on both global context and multi-scale detail from the outset.

\begin{figure*}[!ht]
\centering
\includegraphics[scale=0.054]{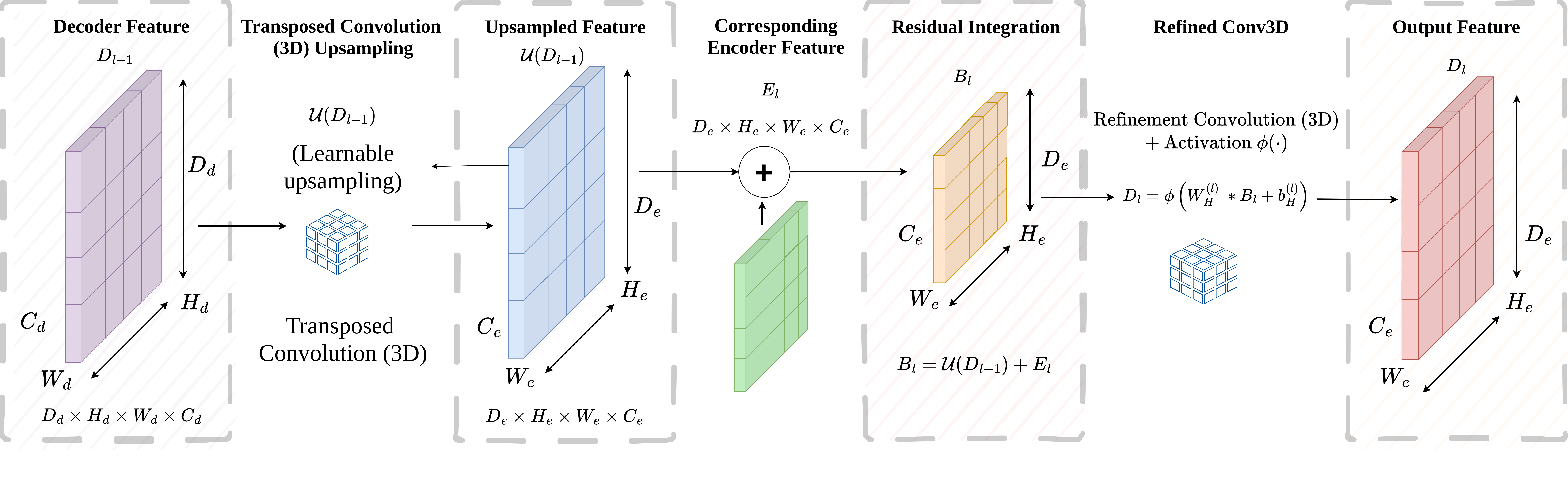}
\caption{Overview of the Hybrid Upsampling and Residual Integration (HURI) module. The decoder feature is upsampled using a learnable transposed convolution, integrated with the corresponding encoder feature via a residual connection, and refined through a final convolution block.}
\label{fig:huri}
\end{figure*}

\paragraph{Hybrid Upsampling and Residual Integration.} Interpolation-based upsampling introduces aliasing and blurs fine tumor boundaries, while transposed convolution alone still cannot recover the high-frequency detail lost during encoding \cite{koc2026review}. At decoder stage $l$, we form the residual bridge following Equation \eqref{eq:B_l}:
\begin{equation}
B_l = \mathcal{U}(D_{l-1}) + E_l
\label{eq:B_l}
\end{equation}
where $\mathcal{U}(\cdot)$ denotes the learned transposed convolution at that stage. The decoder output at stage $l$ is given by:
\begin{equation}
D_l = \phi\big(W_H^{(l)} * B_l + b_H^{(l)}\big)
\label{eq:huri}
\end{equation}
Since $E_l$ enters $B_l$ additively before the refinement convolution, the gradient of the training loss with respect to $D_{l-1}$ retains a direct path through the encoder,
\begin{equation}
\frac{\partial \mathcal{L}}{\partial D_{l-1}} = \frac{\partial \mathcal{L}}{\partial D_l}\left(W_H^{(l)} \frac{\partial \mathcal{U}}{\partial D_{l-1}} + W_H^{(l)} \frac{\partial E_l}{\partial D_{l-1}}\right)
\label{eq:huri_grad}
\end{equation}
which prevents the upsampling path alone from bottlenecking gradient flow to earlier decoder stages, following the same identity-mapping argument used to justify residual learning in general. Figure \ref{fig:huri} visualizes this upsampling and residual integration pipeline.

\paragraph{Residual Skip.} A standard skip connection forwards $E_l$ to the decoder unchanged, which assumes encoder and decoder features are already compatible at every scale. We instead apply a learnable filter to the encoder feature before fusion, giving the transformed residual $\mathcal{R}(E_l) = W_R^{(l)} * E_l + b_R^{(l)}$, merged with the decoder output as $Q_l = D_l + \mathcal{R}(E_l)$. The skip output at stage $l$ is given by:
\begin{equation}
S_l = \phi\big(W_S^{(l)} * Q_l + b_S^{(l)}\big)
\label{eq:rskip}
\end{equation}
Because $\mathcal{R}(E_l)$ is itself learnable rather than fixed, the gradient reaching the encoder scales directly with $W_R^{(l)}$,
\begin{equation}
\frac{\partial \mathcal{L}}{\partial E_l} = \frac{\partial \mathcal{L}}{\partial S_l} \cdot \phi'(P_l^S) \cdot W_S^{(l)} \cdot W_R^{(l)}
\label{eq:rskip_grad}
\end{equation}
which a standard, untransformed skip connection cannot provide, since its contribution to this gradient would be fixed at identity rather than learned. 

Finally, since deeper decoder stages capture coarse tumor shape while shallower stages capture fine boundary detail, we aggregate across all $L$ stages rather than relying on the final stage alone,
\begin{equation}
A = \sum_{l=1}^{L} \gamma_l \, S_l, \qquad \gamma_l = \frac{\exp(w_l)}{\sum_{l'=1}^{L}\exp(w_{l'})}
\label{eq:aggregation}
\end{equation}
where $w_l$ are learnable logits, and the softmax parameterization guarantees $\sum_l \gamma_l = 1$ by construction, rather than as an added constraint on unconstrained weights.

\subsubsection{Coarse Mask Generation}
\label{subsec_coarse_mask_generation}
The radiomic conditioning mechanism requires a tumor mask before the final decoder stage completes, since it must extract descriptors from a region estimate. We obtain this estimate from decoder stage $D_{L-1}$, one stage before the final output, through a projection to class probabilities:
\begin{equation}
\hat{Y}_{\text{coar}} = \sigma\big(W_c \, D_{L-1} + b_c\big)
\label{eq:coarse_mask}
\end{equation}
where $W_c, b_c$ are the learnable weights and bias of this projection, and $\sigma(\cdot)$ is the sigmoid function. We supervise this coarse output directly, using the same Dice-based formulation applied to the final segmentation, as shown in Equation \eqref{eq:coarse_dice}:
\begin{equation}
\mathcal{L}_{\text{coar}} = 1 - \frac{2\sum_{i=1}^{N} \hat{Y}_{\text{coar},i} \, Y_i + \epsilon}{\sum_{i=1}^{N} \hat{Y}_{\text{coar},i} + \sum_{i=1}^{N} Y_i + \epsilon}
\label{eq:coarse_dice}
\end{equation}
where $Y_i$ is the ground truth label at voxel $i$, $N$ is the total number of voxels, and $\epsilon$ is a small constant that prevents division by zero. Direct supervision at this stage ensures $\hat{Y}_{\text{coar}}$ is a meaningful tumor estimate on its own, rather than an arbitrary intermediate activation. The total training objective combines this auxiliary term with the final segmentation loss, following Equation \eqref{eq:total_loss}:
\begin{equation}
\mathcal{L}_{\text{total}} = \mathcal{L}_{\text{final}} + \lambda \, \mathcal{L}_{\text{coar}}
\label{eq:total_loss}
\end{equation}
where $\lambda$ weights the contribution of the coarse supervision. Because $\mathcal{L}_{\text{coar}}$ depends on $D_{L-1}$, and $D_{L-1}$ depends on every encoder and decoder parameter before it, gradients from this term reach the earlier stages of the network directly, rather than through the final loss alone.

\subsubsection{Radiomic Descriptor Extraction and Conditioning}
\label{subsec_radiomic_conditioning}
\paragraph{Descriptor extraction.} From the coarse mask $\hat{Y}_{\text{coar}}$, we extract a multi-dimensional radiomic vector $R$, covering first-order intensity statistics and GLCM and GLRLM texture descriptors. We exclude shape descriptors, since properties such as sphericity and surface area are sensitive to small boundary errors, and $\hat{Y}_{\text{coar}}$ has not yet been refined. This extraction step is not differentiable, so gradients do not flow through $R$ itself; the coarse mask is instead trained through $\mathcal{L}_{\text{coar}}$ in Equation \eqref{eq:coarse_dice}. We normalize $R$ using the mean $\mu$ and standard deviation $\sigma$ of each feature, computed once over the training set,
\begin{equation}
\tilde{R} = \text{diag}(\sigma)^{-1}(R - \mu)
\label{eq:radiomics_norm}
\end{equation}
where $\text{diag}(\sigma)^{-1}$ rescales each of the dimensions independently, since raw radiomic features differ in scale by several orders of magnitude. A projection layer then compresses $\tilde{R}$ into a lower-dimensional embedding,
\begin{equation}
Z = W_p \, \tilde{R} + b_p
\label{eq:radiomics_proj}
\end{equation}
where $W_p, b_p$ are learnable, and $Z$ has substantially fewer dimensions than $\tilde{R}$, reducing redundancy among correlated texture descriptors.

\paragraph{Conditioning mechanism.} We use $Z$ to generate a channel-wise scale and shift for the final backbone representation $A$, rather than concatenating $Z$ directly into the network. The scale and shift terms are produced through two learnable projections,
\begin{equation}
\eta = W_\eta \, Z + b_\eta, \qquad \zeta = W_\zeta \, Z + b_\zeta
\label{eq:film_params}
\end{equation}
and applied to $A$ through an element-wise product and addition,
\begin{equation}
A' = \eta \odot A + \zeta
\label{eq:film_apply}
\end{equation}
where $\odot$ denotes the Hadamard product, broadcast across the spatial dimensions of $A$ so that each channel is scaled and shifted uniformly. Our conditioning signal $Z$ is generated internally from the network's own coarse prediction and requires no ground truth mask at inference. The final segmentation allows the radiomic descriptors extracted from the coarse stage to directly influence the refined output.

\begin{figure*}[!ht]
  \centering
  \includegraphics[scale=0.1]{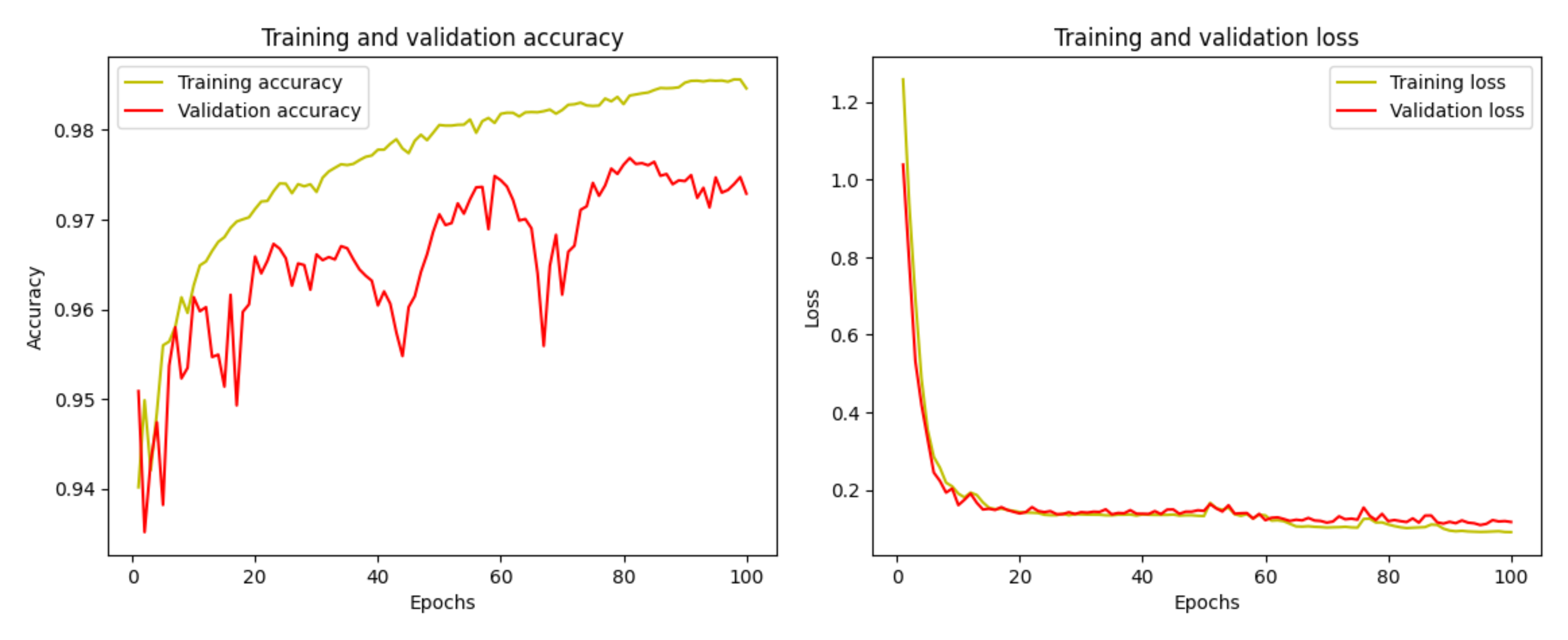}
  \caption{Epoch-wise accuracy and loss for training and validation}
  \label{fig:train_and_val_curve}
\end{figure*}

\begin{figure*}[!ht]
\centering
\includegraphics[scale=0.55]{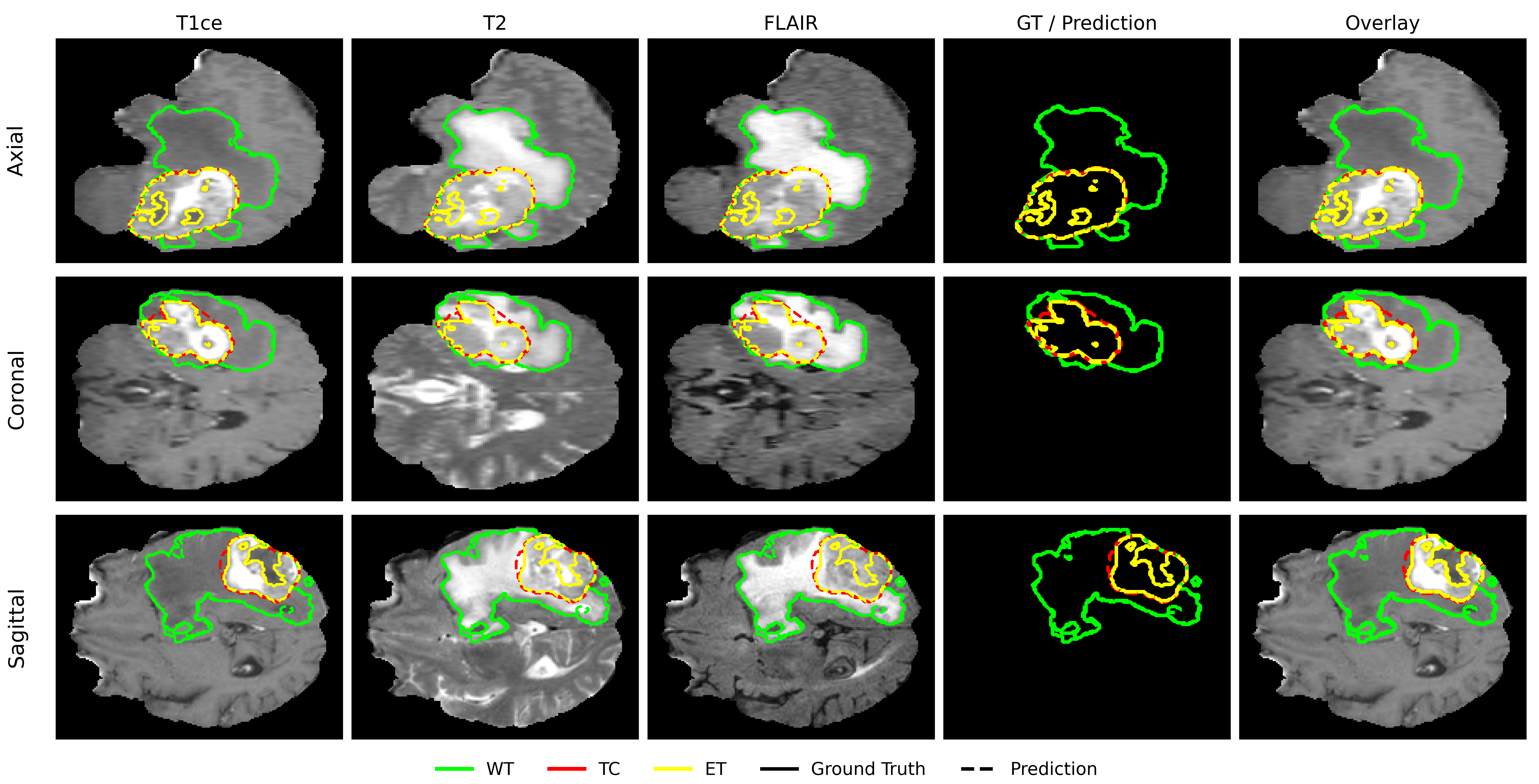}
\caption{Qualitative random slice segmentation on a high-grade glioma (HGG) case. Rows show axial, coronal, and sagittal views; columns show T1ce, T2, and FLAIR modalities with overlaid contours, a contour-only ground-truth/prediction comparison, and an overlay. Solid lines denote ground truth and dashed lines denote model predictions, with WT, TC, and ET shown in green, red, and yellow, respectively.}
\label{fig:hgg_vis}
\vspace{-12pt}
\end{figure*}

\section{Result Analysis}
\label{sec_analysis_of_results}

\subsection{Experimental Setup}
\label{subsec_Experimental_setup}
{The experiments were conducted on a system configured with an AMD Ryzen 5 5600X 6-core CPU, 16 GB of RAM, and a ZOTAC RTX3060 GPU with 12 GB VRAM.} The datasets were processed by converting the NIfTI files into .npy format. Since the official ground truths for BraTS validation and test sets are not publicly available, we used the available set. We split this data into training (75\%) and validation (25\%) subsets. Table \ref{tab:hyperparameters_summary} details the hyperparameters and training setup.

\begin{table}[!ht]
\centering
\caption{Summary of hyperparameters and configuration settings used in ReFRM3D.}
\label{tab:hyperparameters_summary}
\resizebox{\columnwidth}{!}{%
	\begin{tabular}{lll}
	\midrule
	\textbf{Category} & \textbf{Parameter} & \textbf{Value} \\
	\midrule
	{Preprocessing} 
	& Intensity threshold & 99th percentile \\
	& Margin cropping ($\delta$) & 1.5$\sigma$ \\
	& Input size & $128\times128\times128$ \\
	& Input channels & 3 (FLAIR, T1ce, T2) \\
	\midrule
	{Training Config.} 
	& Optimizer & Adam \\
	& Learning rate & 0.0001 \\
	& Batch size & 2 \\
	& Train epochs & 100 \\
	& Data split (train/val) & 75-25\% \\
	& Random seed {(fixed)} & 42 \\
	\midrule
	{Loss Func.} 
	& Loss type & Dice + Focal \\
	& Focal loss gamma ($\gamma$) & 2 \\
	& Class weighting & Applied \\
	\midrule
	{Architecture} 
	& Convolution kernel ($k$) & $3\times3\times3$ \\
	& Output kernel & $1\times1\times1$ \\
	& Dilation rate ($r$) & 1 \\
	& Stride (convolution) & 1 \\
	& Stride (pool/upsample) & 2 \\
	\midrule
	{Regularization} 
	& Data augmentation & None \\
	& Early stopping & No \\
    & Enc (dropout) & [0.1, 0.1, 0.2, 0.2, 0.3]\\
    & Dec (dropout) & [0.2, 0.2, 0.1, 0.1]\\
	\midrule
	\end{tabular}	
}
\vspace{-5pt}
\end{table}


During model training (see the progression in Figure \ref{fig:train_and_val_curve}), the peak GPU memory usage was approximately 10,851 MB (10.85 GB). Most of the preprocessing operated on the CPU, with only a limited number of data processing steps utilizing the GPU (see Section \ref{preproc_slice}).

\subsection{Training Evaluation} 
\label{subsec_training_evaluation}
Two loss functions were employed while training the model to optimize performance: Dice Loss and Categorical Focal Loss. The Dice Loss was used to address class imbalance by incorporating class weights for each tumor region. It was calculated as shown in Equation \eqref{eq:dice_loss}: 
\begin{equation}
    \mathcal{L}_{\text{Dice}} = 1 - \frac{2 \sum_{i=1}^{N} w_c y_i \hat{y}_i}{\sum_{i=1}^{N} w_c y_i + \sum_{i=1}^{N} w_c \hat{y}_i}
    \label{eq:dice_loss}
\end{equation}
where \(y_i\) represents the ground truth binary mask, \(\hat{y}_i\) denotes the predicted mask, and \(w_c\) is the weight for class \(c\). In addition, we introduced Categorical Focal Loss, formulated as:  
\begin{equation}
    \mathcal{L}_{\text{Focal}} = -\alpha_c (1 - p_t)^\gamma \log(p_t)
    \label{eq:focal_loss}
\end{equation}
where \(\alpha_c\) is the weighting factor for class \(c\), \(p_t\) is the predicted probability for the target class, and \(\gamma\) is the focusing parameter to down-weight easy examples. The total loss function for our model is the summation of the two losses, as defined in Equations \eqref{eq:dice_loss} and \eqref{eq:focal_loss}.

\subsection{Segmentation Evaluation} 
\label{subsec_segmentation_evaluation}
\begin{figure*}[!ht]
\centering
\includegraphics[scale=0.55]{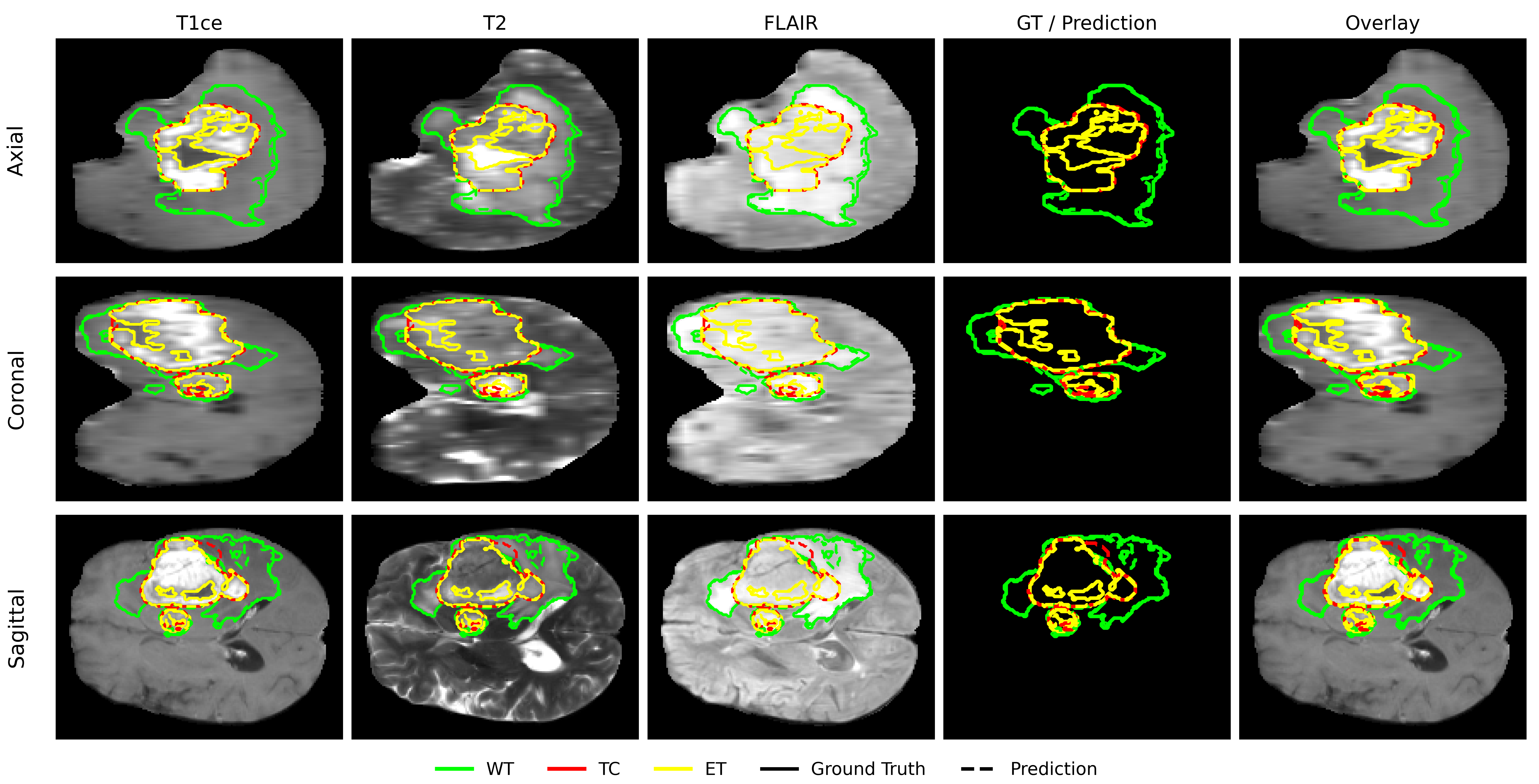}
\caption{Visualization of a random slice of a low-grade glioma (LGG) case. Unlike the HGG case, the LGG lesions are more homogeneous with minimal enhancement.}
\label{fig:lgg_vis}
\end{figure*}

Across all datasets, the T2 modality consistently provided higher segmentation accuracy than T1ce and Flair. In BraTS2019, the T2 modality achieved the highest DSC of 94.59\%, followed by Flair and T1ce. T2 provided the best results in BraTS2020. In terms of mean values, T2 consistently achieved the highest mean DSC (94.92\%) and JCS (90.33\%). Figures \ref{fig:hgg_vis} and \ref{fig:lgg_vis} visualize the high-grade and low-grade glioma subtype segmentation, respectively.

\begin{table*}[!ht]
\small
\centering
\caption{Segmentation results for different modalities. The columns \textbf{\( m_m \, \text{DSC} \)} and \textbf{\( m_m \, \text{JCS} \)} denote the \textbf{m}ean \textbf{D}ice \textbf{S}imilarity \textbf{C}oefficient and \textbf{J}accard \textbf{C}oefficient \textbf{S}imilarity for the respective \textbf{m}odalities.}
\vspace{-5pt}
\label{tab:segmentation_results_modality}
\renewcommand{\arraystretch}{1.1}
	\begin{tabular}{llcclccclc}
	\midrule
	\multirow{2}{*}{} & \multirow{2}{*}{\textbf{Modalities}} & \multicolumn{3}{c}{\textbf{DSC(\%)}} & \multirow{2}{*}{\textbf{$m_m \, \text{DSC}(\%)$}} & \multicolumn{3}{c}{\textbf{JCS(\%)}} & \multirow{2}{*}{\textbf{$m_m \, \text{JCS}(\%)$}} \\ \cmidrule(lr){3-5} \cmidrule(lr){7-9}
	&                           & \textbf{WT}     & \textbf{ET}    & \textbf{TC}    &                                & \textbf{WT}     & \textbf{ET}    & \textbf{TC}    &                                \\ \midrule
	{BraTS2019} & T1ce                  & 92.98 & 91.17 & 92.54 & 92.23   & 86.88 & 83.77 & 85.59 & 85.41   \\
	& T2                                & 95.05 & 94.00 & 94.71 & 94.59   & 90.57 & 88.68 & 89.73 & 89.66   \\
	& Flair                             & 94.10 & 92.87 & 93.66 & 93.54   & 88.86 & 86.69 & 87.88 & 87.81   \\ 
    & \textit{{avg.}}                     & {94.04} & {92.68} & {93.64 $\rightarrow$} & {93.45} & {88.77} & {86.38} & {87.73 $\rightarrow$} & {87.63}   \\
	{BraTS2020} & T1ce                  & 93.30 & 90.25 & 92.74 & 92.09   & 87.44 & 82.23 & 85.38 & 85.02   \\
	& T2                                & 94.78 & 95.08 & 94.90 & 94.92   & 90.08 & 90.62 & 90.33 & 90.34   \\
	& Flair                             & 94.18 & 93.41 & 93.88 & 93.82   & 89.00 & 87.63 & 88.37 & 88.33   \\ 
    & \textit{{avg.}}                     & {94.09} & {92.91} & {93.84 $\rightarrow$} & {93.61} & {88.84} & {86.83} & {88.03 $\rightarrow$} & {87.90}   \\
	{BraTS2021} & T1ce                  & 93.01 & 89.91 & 91.93 & 91.62   & 86.93 & 81.67 & 84.56 & 84.39   \\
	& T2                                & 94.20 & 91.12 & 92.07 & 92.46   & 89.04 & 83.69 & 86.01 & 86.25   \\
	& Flair                             & 93.91 & 90.05 & 92.38 & 92.11   & 88.52 & 81.90 & 85.42 & 85.28   \\ 
    & \textit{{avg.}}                     & {93.70} & {90.36} & {92.13 $\rightarrow$} & {92.06} & {88.16} & {82.42} & {85.33 $\rightarrow$} & {85.30}   \\
    \midrule
	\end{tabular}
\vspace{-7pt}
\end{table*}

Table \ref{tab:segmentation_results_modality} presents the detailed segmentation results for different modalities.

In the BraTS2019 dataset, the model achieved a mean DSC of 93.45\% and a JCS of 87.63\%. As shown in Table \ref{tab:segmentation_comparison_brats2019}, our framework achieves higher DSC values than several transformer-based architectures, including TransBTS, SwinBTS, and UNETR, as well as recent CNN-based methods such as DMRA U-Net and BC-TSEA-UNet. Compared with the five best-performing models listed in the table, our model achieves improvements of +3.68\%, +6.42\%, and +6.06\% for WT, ET, and TC, respectively, with an average DSC improvement of +5.79\%.

\begin{table}[ht]
\centering
\small
\caption{Comparison of segmentation performance on BraTS2019 dataset. {The best and the second-best results are \textbf{bolded} and \underline{underlined}, respectively.}}
\vspace{-5pt}
\label{tab:segmentation_comparison_brats2019}
\begin{tabular}{lcccc}
\midrule
\textbf{Ref}               & \textbf{WT}  & \textbf{ET} & \textbf{TC} & \textbf{Avg.} \\
\midrule
Oktay et al. \cite{oktay2018attention}  & {0.888} & {0.759} & {0.772} & {0.806} \\
Wang et al. \cite{wang2021transbts}    & {0.900} & {0.789} & {0.819} & {0.836} \\
Valanarasu et al. \cite{valanarasu2021kiu}   & {0.876} & {0.732} & {0.739} & {0.782} \\
Barzegar et al. \cite{barzegar2021wlfs}    & 0.901 & \underline{0.890} & \underline{0.887} & \underline{0.893} \\
Ullah et al. \cite{ullah2022cascade}    & 0.899 & 0.845 & 0.845 & 0.863 \\
Li et al. \cite{li2022automatic}     & 0.890 & 0.835 & 0.842 & 0.856 \\
Xu et al. \cite{xu2022brain}         & 0.900 & 0.740 & 0.820 & 0.820 \\
Jiang et al. \cite{jiang2022swinbts}    & {0.897} & {0.744} & {0.792} & {0.811} \\
Hatamizadeh et al. \cite{hatamizadeh2022unetr}& {0.885} & {0.672} & {0.744} & {0.767} \\
Liu et al. \cite{liu2023attention}    & 0.892 & 0.767 & 0.811 & 0.823 \\
Tong et al. \cite{tong2023dual}        & 0.885 & 0.751 & 0.776 & 0.804 \\
Jawad et al. \cite{jawad2023gsnet}      & {0.897} & {0.790} & {0.869} & {0.852} \\
Liu et al. \cite{liu2023multiscale}   & {0.899} & {0.779} & {0.839} & {0.839} \\
Zhang et al. \cite{zhang2024dilated}    & {0.901} & {0.881} & {0.886} & {0.889} \\
Schwehr et al. \cite{schwehr2025brain}    & {0.844} & {0.827} & {0.865} & {0.845} \\
Zhang et al. \cite{zhang2025light}      & {0.888} & {0.862} & {0.868} & {0.873} \\
Zhou et al. \cite{zhou2025comprehensive} & {0.900} & {0.826} & {0.799} & {0.842} \\
Peng et al. \cite{peng2026adml}        & {--} & {--} & {--} & {0.847} \\
Jadhav et al. \cite{jadhav2026bias}      & {\underline{0.916}} & {0.810} & {0.869} & {0.865} \\
\rowcolor{gray!20} ReFRM3D (Ours) & \textbf{0.940} & \textbf{0.926} & \textbf{0.936} & \textbf{0.934} \\
\midrule
\end{tabular}
\vspace{-5pt}
\end{table}

For BraTS2020, the model recorded a mean DSC of 93.61\% and a JCS of 87.90\%. Table \ref{tab:segmentation_comparison_brats2020} shows that ReFRM3D surpasses baselines including nnU-Net, TransBTS, and UNETR. Furthermore, when compared with the five strongest competing models for each subregion, our method achieves average DSC gains of +2.01\% in WT, +11.31\% in ET, and +5.62\% in TC, corresponding to an overall average improvement of +7.25\%.

\begin{table}[!ht]
\centering
\small
\caption{Comparison of segmentation performance on BraTS2020 dataset. {The best and the second-best results are \textbf{bolded} and \underline{underlined}, respectively.}}
\label{tab:segmentation_comparison_brats2020}
\begin{tabular}{lcccc}
\midrule
\textbf{Ref} & \textbf{WT}  & \textbf{ET} & \textbf{TC} & \textbf{Avg.} \\
\midrule
Oktay et al. \cite{oktay2018attention}      & {0.855} & {0.718} & {0.759} & {0.777} \\ 
Qamar et al. \cite{qamar2020hi}             & {0.874} & {0.794} & {0.837} & {0.835} \\ 
Isensee et al. \cite{isensee2021nnu}        & 0.850  & {0.820}  & 0.895 & 0.855 \\
Wang et al. \cite{wang2021transbts}        & {0.900} & {0.787} & {0.817} & {0.835} \\ 
Xu et al. \cite{xu2022brain}             & 0.900  & 0.780 & 0.820   & 0.833 \\
Jiang et al. \cite{jiang2022swinbts}        & {0.890} & {0.773} & {0.803} & {0.822} \\ 
Hatamizadeh et al. \cite{hatamizadeh2022unetr}    & {0.883} & {0.711} & {0.758} & {0.784} \\ 
Liu et al. \cite{liu2023multiscale}       & 0.899  & 0.780   & 0.883 & 0.854 \\
Liu et al. \cite{liu2023attention}        & {0.883} & {0.780} & {0.817}  & {0.827} \\
Raza et al. \cite{raza2023dresu}           & {0.866} & {0.800} & {0.835}  & {0.834} \\ 
Jawad et al. \cite{jawad2023gsnet}          & {0.924}  & {0.814} & {\underline{0.910}} & {\underline{0.883}} \\ 
Domadia et al. \cite{domadia2024segmenting}   & 0.917 & 0.758 & {0.769} & 0.815 \\
Zhou et al. \cite{zhou2025comprehensive}   & {0.909} & {0.802} & {0.827} & {0.846} \\ 
Jadhav et al. \cite{jadhav2026bias}          & {\underline{0.936}} & {\underline{0.844}} & {0.859} & {0.880} \\ 
He et al. \cite{he2026towards}           & {0.918} & {0.726} & {0.864} & {0.836} \\
\rowcolor{gray!20} ReFRM3D (Ours) &  \textbf{0.940} &  \textbf{0.929} &  \textbf{0.938} & \textbf{0.936} \\
\midrule
\end{tabular}
\vspace{-10pt}
\end{table}

In BraTS2021, the mean DSC was 92.06\%, and the JCS was 85.30\%. Table \ref{tab:segmentation_comparison_brats2021} indicates that while some recent methods achieve competitive tumor core performance, our approach provides balanced segmentation across all subregions. In this dataset, compared to the top five models for each tumor subregion, our model achieves improvements of +0.84\% for WT, +2.18\% for ET, and +1.31\% for TC, with an overall average gain of +0.96\% relative to their mean performance.

\begin{table}[!ht]
\centering
\caption{Comparison of segmentation performance on BraTS2021 dataset. {The best and the second-best results are \textbf{bolded} and \underline{underlined}, respectively.}}
\label{tab:segmentation_comparison_brats2021}
\resizebox{\columnwidth}{!}{%
\begin{tabular}{lcccc}
\midrule
\textbf{Ref} & \textbf{WT}  & \textbf{ET} & \textbf{TC} & \textbf{Avg.} \\
\midrule
Hatamizadeh et al. \cite{hatamizadeh2021swin}   & {0.927} & {0.868} & {0.899} & {0.898} \\ 
Isensee et al. \cite{isensee2021nnu} & {--}  & {--} & {--} & {0.912} \\ 
Jiang et al. \cite{jiang2022swinbts}      & {0.918} & {0.832} & {0.847} & {0.866} \\ 
Li et al. \cite{li2023transu2}         & {0.923} & {0.858} & {0.863} & {0.881} \\ 
Lin et al. \cite{lin2023ckd}            & {0.933} & {0.885} & {0.901} & {0.906} \\ 
Roy et al. \cite{roy2023mednext} & {--}  & {--} & {--} & {\underline{0.915}} \\
Domadia et al. \cite{domadia2024segmenting} & 0.902 & 0.758 & 0.812 & 0.824 \\
Sun et al. \cite{sun2024glioma}         & 0.889 & --   & 0.884 & -- \\
Li et al. \cite{li2024multi}           & \textbf{0.941} & {0.871} & 0.911 & 0.908 \\
Isensee et al. \cite{isensee2024nnu} & {--}  & {--} & {--} & {0.912} \\
Schwehr et al. \cite{schwehr2025brain}      & {0.905} & {\underline{0.897}} & {\textbf{0.921}} & {0.908} \\ 
Zhou et al. \cite{zhou2025comprehensive} & {0.919} & {0.867} & {0.899} & {0.895} \\ 
Asimeng et al. \cite{asimeng2026dynamic} & {0.917} & {0.888} & {0.909} & {0.905} \\
\rowcolor{gray!20} ReFRM3D (Ours)               &  \underline{0.937} &  \textbf{0.903} &  \textbf{0.921} & {\textbf{0.920}}\\
\midrule
\end{tabular}}
\end{table}

\begin{table}[!ht]
\centering
\small
\caption{{Classification performance alongside recent methods on the BraTS benchmark, reported for context only. (--) denotes: values not given.}}
\label{tab:classification_comparison}
\begin{tabular}{lcccc}
\midrule
\textbf{Ref} & \textbf{Dataset} & \textbf{Acc.} & \textbf{Sen} & \textbf{Spec} \\
\midrule
\cite{zhou2025comprehensive}  & BraTS19 & -- & 75.84\% & 76.11\% \\
\cite{jabbar2023brain}        & BraTS19 & 98.90\% & 98.70\% & -- \\
\cite{jadhav2026bias}         & BraTS19 & 99.09\% & 99.15\% & 99.98\% \\
\cite{feng2025efficient}      & BraTS19-21 & 98.10\% & 99.05\% & 95.20\% \\
\cite{jabbar2023brain}        & BraTS20 & 99.60\% & 99.00\% & -- \\
\cite{jadhav2026bias}         & BraTS20 & 99.88\% & 99.50\% & 99.97\% \\
\cite{salama2026brain}        & BraTS21 & 95.24\% & 95.81\% & -- \\
\cite{mittal2026enhancing}    & BraTS23 & 90.85\% & 91.01\% & 98.00\% \\
\cite{wang2025comparative}    & BraTS24 & 88.00\% & 86.00\% & 89.00\% \\
\cite{jadhav2026bias}         & BraTS24 & 99.82\% & 98.50\% & 99.49\% \\
\midrule
Ours                          & BraTS19 & 99.27\% & 99.33\% & 99.20\% \\
                              & BraTS20 & 98.83\% & 98.57\% & 98.87\% \\
                              & BraTS21 & 98.70\% & 98.53\% & 99.07\% \\
\midrule
\end{tabular}
\vspace{-10pt}
\end{table}
\subsection{Auxiliary Classification}
\label{subsec_classification_evaluation}
As a secondary output, the radiomic embedding $Z$ (see Section \ref{subsec_radiomic_conditioning}) additionally feeds a lightweight classification head for subregion label prediction.

\begin{sidewaystable}
\caption{{Classification results for different tumor sub-regions.}}
\label{tab:classification_results_subregions}
\renewcommand{\arraystretch}{3}

\begin{tabular*}{\textheight}{@{\extracolsep\fill}l p{0.85cm} p{0.85cm} p{0.85cm} p{0.85cm} p{0.85cm} p{0.85cm} p{0.85cm} p{0.85cm} p{0.85cm} p{0.85cm} p{0.85cm} p{0.85cm} p{0.85cm} p{0.85cm} p{0.85cm} p{0.85cm}@{}}
\midrule
\multirow{2}{*}{Dataset} & \multicolumn{4}{@{}c@{}}{Accuracy (\%)} & \multicolumn{4}{@{}c@{}}{Precision (\%)} & \multicolumn{4}{@{}c@{}}{Sensitivity/Recall (\%)} & \multicolumn{4}{@{}c@{}}{Specificity (\%)} \\ \cmidrule(lr){2-5} \cmidrule(lr){6-9} \cmidrule(lr){10-13} \cmidrule(lr){14-17}
            & WT    & TC    & ET    & \cellcolor{gray!20} Avg.   & WT    & TC    & ET    & \cellcolor{gray!20} Avg.   & WT    & TC    & ET    & \cellcolor{gray!20} Avg.   & WT    & TC    & ET    & \cellcolor{gray!20} Avg.   \\ \midrule
BraTS2019   & 99.7  & 99.3  & 98.8  & \cellcolor{gray!20} 99.27  & 99.0  & 99.1  & 98.4  & \cellcolor{gray!20} 98.83  & 99.3  & 99.8  & 98.9  & \cellcolor{gray!20} 99.33  & 99.4  & 99.2  & 99.0  & \cellcolor{gray!20} 99.20  \\
BraTS2020   & 99.4  & 99.3  & 97.8  & \cellcolor{gray!20} 98.83  & 99.5  & 99.1  & 98.2  & \cellcolor{gray!20} 98.93  & 98.9  & 99.7  & 97.1  & \cellcolor{gray!20} 98.57  & 99.8  & 99.8  & 97.0  & \cellcolor{gray!20} 98.87  \\
BraTS2021   & 99.2  & 98.9  & 98.0  & \cellcolor{gray!20} 98.70  & 99.7  & 99.4  & 97.7  & \cellcolor{gray!20} 98.93  & 99.1  & 98.1  & 98.5  & \cellcolor{gray!20} 98.57  & 99.4  & 99.1  & 98.7  & \cellcolor{gray!20} 99.07  \\ \midrule
\end{tabular*}
\end{sidewaystable}

Table \ref{tab:classification_comparison} compares these findings with recent classification-focused methods. As seen in Table \ref{tab:classification_results_subregions}, the label prediction accuracy remains consistently above 97\% across all subregions and datasets, with ET showing the lowest values. 

This follows the same pattern seen in segmentation: ET is the smallest and innermost subregion under the BraTS label hierarchy. Consequently, it carries the fewest voxels and is most sensitive to upstream segmentation. WT, being the largest and most spatially stable region, is least affected. 

\subsection{Comparison with Existing Approaches and Generalizability}
Existing methods employ various architectural enhancements, such as multi-scale feature learning \cite{ding2020multi, zhou20203d}, attention mechanisms \cite{liu2023multiscale}, and feature reuse modules \cite{xue2020hypergraph, zhou2021erv}, to improve tumor segmentation.

Our model achieves higher Dice scores across all tumor subtypes for segmentation. Notably, on BraTS2020, we obtain a DSC of 94\%, 92.9\%, and 93.8\% for WT, ET, and TC. Table \ref{tab:dsc_comparison} details the comparison.

\begin{table}[!ht]
\centering
\caption{Comparison of Dice (\%) for different datasets and methods. The best results are highlighted in bold, while the second-best results are underlined.}
\label{tab:dsc_comparison}
\resizebox{\columnwidth}{!}{%
\begin{tabular}{lcccc} 
\midrule
\multirow{2}{*}{\textbf{Model}} & \multirow{2}{*}{\textbf{BraTS}} & \multicolumn{3}{c}{Dice (\%)} \\ \cmidrule(lr){3-5} 
& & WT & ET & TC  \\ \midrule
DDU-net \cite{jiang2021novel}        & 2017 & 0.838 & 0.783 & 0.898\\
DenseNets \cite{ding2020multi}         & 2015 & 0.816 & 0.818 & 0.858 \\
U-Net \cite{li2019novel}           & 2015 & 0.733 & 0.726 & 0.890 \\
                                & 2017 & 0.763 & 0.642 & 0.876 \\
C-ConvNet \cite{ranjbarzadeh2021brain} & 2018 & 0.872 & 0.911 & 0.920 \\
MSMANet \cite{zhang2021msmanet}      & 2018 & 0.890 & 0.758 & 0.811 \\
F 2 FCN \cite{xue2020hypergraph}     & 2013 & 0.780  & 0.750  & 0.880 \\
                                     & 2015 & 0.870  & 0.800   & 0.750 \\
ERV-Net \cite{zhou2021erv}           & 2018 & 0.866 & 0.818 & 0.912 \\
DenseAFPNet \cite{zhou20203d}            & 2013 & 0.720  & 0.700   & 0.870  \\
                                         & 2015 & 0.700   & 0.620  & 0.830  \\
                                         & 2018 & 0.773 & 0.752 & 0.864 \\
WLFS \cite{barzegar2021wlfs}      & 2015 & 0.894 & 0.901 & 0.915 \\
                                  & 2017 & 0.872 & 0.886 & 0.900   \\
CH-UNet \cite{xu2022brain}           & 2018 & 0.850  & 0.790  & 0.910  \\
2D-CNNs \cite{zhang20213d}           & 2018 & 0.850  & 0.830  & 0.820  \\
                                     & 2013 & 0.780 & 0.830  & 0.810  \\
(TS)\(^2\)WM \cite{zhong20202wm}          & 2017 & 0.700   & 0.760  & 0.890  \\
\midrule
{ReFRM3D (Ours)} & 2019 & \underline{0.940} & \underline{0.926} & \underline{0.936} \\
& 2020 & \textbf{0.940} & \textbf{0.929} & \textbf{0.938} \\
& 2021 & 0.937 & 0.903 & 0.921 \\
\midrule
\end{tabular}}
\end{table}

{To evaluate cross-dataset generalization, we evaluated the model on the BraTS-Africa dataset \cite{adewole2025brats}. The dataset differs from BraTS2019--2021 in terms of scanner field strength (1.5T), acquisition conditions, and patient demographics.}

\begin{table}[!ht]
\centering
\small
\caption{{Comparison of segmentation performance on the BraTS-Africa dataset.}}
\label{tab:generalizability_comparison}
\begin{tabular}{lcccc}
\midrule
\textbf{Method} & \textbf{ET} & \textbf{TC} & \textbf{WT}  & \textbf{Avg.} \\
\midrule
nnU-Net         & 0.850 & 0.853 & 0.913 & 0.872 \\
MedNeXt Ens.    & 0.878 & 0.876 & 0.924 & 0.893 \\
BrainUNet       & 0.759 & 0.791 & 0.869 & 0.806 \\
\rowcolor{gray!20} ReFRM3D (Ours)  & 0.843 & 0.880 & 0.911 & 0.878 \\
\midrule
\end{tabular}
\vspace{-6pt}
\end{table}

\begin{table}[!ht]
\centering
\small
\caption{{Statistical analysis of ReFRM3D segmentation performance on the BraTS-Africa dataset ($n = 146$). The 95\% confidence interval (CI) was estimated from three independent runs to assess result variability.}}
\label{tab:generalizability_stats}
\begin{tabular}{lcc}
\midrule
\textbf{Metric} & \textbf{Mean $\pm$ Std} & \textbf{95\% CI} \\
\midrule
ET  & $0.843 \pm 0.047$ & $[0.832,\ 0.854]$ \\
TC  & $0.880 \pm 0.035$ & $[0.872,\ 0.888]$ \\
WT  & $0.911 \pm 0.028$ & $[0.905,\ 0.917]$ \\
Avg & $0.878 \pm 0.037$ & $[0.869,\ 0.887]$ \\
\midrule
\end{tabular}
\vspace{-7pt}
\end{table}

As reported in Table \ref{tab:generalizability_comparison}, ReFRM3D achieved an average DSC score of 0.878, compared with 0.872 for nnU-Net \cite{isensee2021nnu}, 0.806 for BrainUNet \cite{sidume2025resource}, and 0.893 for the MedNeXt Ensemble \cite{roy2023mednext}.

\begin{figure}[ht]
\centering
\includegraphics[scale=0.55]{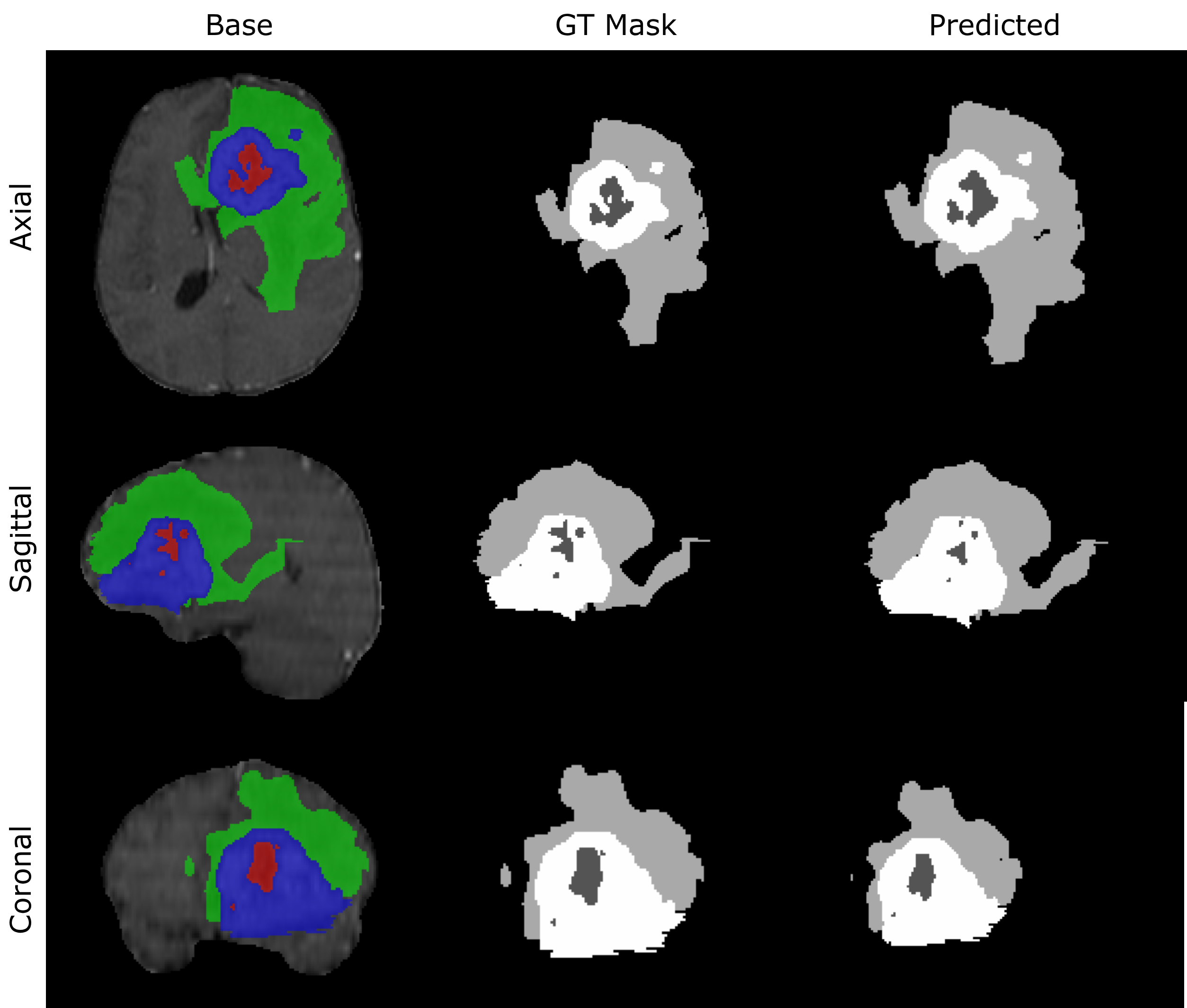}
\caption{Visualization of randomly selected slices for three tumor subtypes in the BraTS-Africa dataset.}
\label{fig:bratsafrica}
\end{figure}

{To further characterize the reliability of these results, we also report the per-subregion mean, standard deviation, and 95\% confidence intervals (CIs) across all test cases. As seen in Table \ref{tab:generalizability_stats}, the standard deviation values were reasonably low across all subregions. Our model also generalizes to a cross-dataset with different imaging characteristics and patient demographics (Figure \ref{fig:bratsafrica}).}

\vspace{-5pt}
\subsection{Model Complexity Analysis}
\label{subsec_complexity}
Table \ref{tab:complexity_comparison} compares the computational complexity of ReFRM3D with several state-of-the-art models. With approximately 8.64M parameters, ReFRM3D has a substantially smaller parameter count than the compared methods. This efficiency is enabled by its lightweight encoder-decoder architecture and the use of dilated convolutions to capture broader contextual information without the additional computational burden of heavy self-attention.

\begin{table*}[!ht]
\centering
\small
\caption{{Complexity comparison with the SOTA models. Our FLOPs are reported for the input size $128\times128\times128$.}}
\label{tab:complexity_comparison}
\begin{tabular}{lllll}
\midrule
\textbf{Ref.} & \textbf{Model} & \textbf{Features} & \textbf{Params (M)} & \textbf{FLOPs (G)}\\
\midrule
\cite{isensee2021nnu}       & nnUNet         & CNN                        & 19.07 & 412.65 \\
\cite{hatamizadeh2021swin}  & SwinUNETR      & Swin Transformer           & 61.98 & 394.84 \\
\cite{chen2021transunet}    & TransUNet      & Hybrid CNN‑Transformer     & 96.07 & 48.34  \\
\cite{amit2021segdiff}      & SegDiff        & U‑Net + Diffusion          & 23    & 82.1   \\
\cite{hatamizadeh2022unetr} & UNETR          & Vision Transformer         & 92.58 & 41.19  \\
\cite{wu2024medsegdiff}     & MedSegDiff     & U‑Net + Diffusion          & 25    & 1770   \\
\cite{wu2024medsegdiffv2}   & MedSegDiff-V2  & U‑Net + Diffusion          & 46    & 983    \\
\cite{sidume2025resource}   & BrainUNet      & U‑Net + Residual Attention & 22    & --     \\
\rowcolor{gray!20} Ours & ReFRM3D & U‑Net + FMFF + HURI + rSkip & 8.64 & 263.5 \\
\midrule
\end{tabular}
\end{table*}

To further characterize the computational and memory requirements of ReFRM3D, Table \ref{tab:complexity_system} reports its overall system complexity. For an input volume of $128\times128\times128$, the model requires 263.54G FLOPs, corresponding to approximately 30,490 FLOPs per parameter. Of the 8.64M parameters, only 4,896 are non-trainable BatchNorm statistics. The FP32 Keras HDF5 weights occupy 68.0 MB, while FP16 storage reduces the inference memory footprint to 16.48 MB. 

A detailed distribution of the model parameters across its components is provided in Table \ref{tab:complexity_breakdown}.

\begin{table}[ht]
\centering
\caption{{System-level complexity of the proposed model.}}
\label{tab:complexity_system}
\resizebox{\columnwidth}{!}{%
\begin{tabular}{ll}
\midrule
\textbf{Metric} & \textbf{$\sim$Value} \\
\midrule
Total Params         & 8.64 M \\
Trainable Params     & 8.64 M \\
Non-trainable Params & 4.9 K (BatchNorm stats) \\
FLOPs ($128^3$ input) & 263.54 GFLOPs \\
FLOPs per Param      & 30.49 K \\
\midrule
Mem: weights (FP32, .h5)      & 68.0 MB \\
Mem: weights (FP16 inference) & 16.48 MB \\
\midrule
\end{tabular}
}
\end{table}

\begin{table}[!ht]
\centering
\small
\caption{{Parameter distribution by architectural component.}}
\label{tab:complexity_breakdown}
\begin{tabular}{ll}
\midrule
\textbf{Component} & \textbf{Parameters} \\
\midrule
Encoder (L1–L4)              & 880,288   \\
Bottleneck + Context Pathway & 4,426,768 \\
FMFF (1×1 alignment convs)   & 22,000    \\
Decoder L4 (HURI + rSkip + RadCond.)    & 2,491,792 \\
Decoder L3 (HURI + rSkip + RadCond.)    & 623,744   \\
Decoder L2 (HURI + rSkip + RadCond.)    & 156,352   \\
Decoder L1 (HURI + rSkip + RadCond.)    & 35,300    \\
Aggregation + Output head    & 596       \\
Aux. classification head    & 6,348     \\
\midrule
{Total} & {8,643,188} \\
\midrule
\end{tabular}
\vspace{-7pt}
\end{table}

\begin{table*}[!ht]
\centering
\scriptsize
\renewcommand{\arraystretch}{1.2}
\caption{{Component-wise ablation study on BraTS2021 (100 randomly selected samples from both HGG and LGG grades, trained for 50 epochs). Each component group is evaluated to isolate its individual contribution. $\Delta$Mean is relative to the baseline of each component group.}}
\label{tab:component_ablation}
\begin{tabular}{l|c|l|ccccc}
\midrule
\multirow{2}{*}{\textbf{Com.}} &
\multirow{2}{*}{\textbf{Exp.}} &
\multirow{2}{*}{\textbf{Variant}} &
\multicolumn{3}{c}{\textbf{DSC (\%)}} &
\multirow{2}{*}{\textbf{Mean}} &
\multirow{2}{*}{$\boldsymbol{\Delta}$\textbf{Mean}} \\
\cline{4-6}
& & & \textbf{WT}  & \textbf{ET} & \textbf{TC} & & \\
\midrule

Fusion
& F1 & No fusion                      & 76.3 & 72.1 & 74.8 & 74.40 & 0.00 \\
& F2 & Single-level fusion            & 78.5 & 74.3 & 76.9 & 76.57 & $+$2.17 \\
& F3 & w/o scale alignment            & 80.1 & 76.2 & 78.7 & 78.33 & $+$3.93 \\
& F4 & w/o context pathway            & 81.4 & 77.8 & 80.1 & 79.77 & $+$5.37 \\
& F5 & + Summation+1$\times$1 (FMFF)   & {82.7} & {79.3} & {81.5} & {81.17} & $+${6.77} \\
\midrule

Upsample
& U1 & Trilinear only                 & 78.2 & 74.6 & 77.1 & 76.63 & 0.00 \\
& U2 & Transposed conv only           & 80.4 & 76.8 & 79.3 & 78.83 & $+$2.20 \\
& U3 & w/o residual integration       & 81.3 & 77.9 & 80.4 & 79.87 & $+$3.24 \\
& U4 & Trans+Direct residual          & 82.8 & 79.4 & 81.7 & 81.30 & $+$4.67 \\
& U5 & Trans+Res+Refine (HURI)        & {83.6} & {80.2} & {82.5} & {82.10} & $+${5.47} \\
\midrule

Skip
& S1 & No skip                 & 76.8 & 72.9 & 75.4 & 75.03 & 0.00 \\
& S2 & Standard direct copy    & 80.2 & 76.5 & 78.9 & 78.53 & $+$3.50 \\
& S3 & Dense skip              & 81.7 & 77.8 & 80.3 & 79.93 & $+$4.90 \\
& S4 & w/o learnable filter    & 83.1 & 79.3 & 81.8 & 81.40 & $+$6.37 \\
& S5 & Conv+Residual, rSkip    & {84.0} & {80.6} & {82.1} & {82.23} & $+${7.20} \\
\midrule

RadCond.
& R1 & Stripped baseline, w/o backbone \& radiomics & 73.1 & 68.8 & 71.5 & 71.14 & 0.00 \\
& R2 & Backbone only (FMFF+HURI+rSkip), single-pass & 86.4 & 82.2 & 83.8 & 84.13 & $+$12.99 \\
& R3 & Backbone + coarse-to-fine, w/o conditioning   & 88.1 & 85.3 & 87.6 & 87.00 & $+$15.86 \\
& R4 & Backbone + coarse-to-fine + radiomic cond.   & {89.5} & {86.9} & {88.0} & {88.13} & $+${16.99} \\
\midrule
\end{tabular}
\vspace{-5pt}
\end{table*}

\begin{table*}[!ht]
\centering
\small
\caption{Hyperparameter configurations and performance metrics. The columns \textbf{LR} and \textbf{T/E (s)} represent the \textbf{L}earning \textbf{R}ate and the average \textbf{T}ime per \textbf{E}poch in seconds, respectively. {This setting uses the Adam optimizer with the input size of $128\times128\times128$}.}
\label{tab:hyperparameter_configurations}
\begin{tabular}{llcccccccc}
\midrule
\multirow{2}{*}{\textbf{Dataset}} & \multirow{2}{*}{\textbf{LR}} & \multirow{2}{*}{\textbf{Loss}} & \multicolumn{2}{c}{\textbf{Training}} & \multicolumn{2}{c}{\textbf{Validation}} & \multirow{2}{*}{\textbf{T/E (s)}} \\
\cmidrule(lr){4-5} \cmidrule(lr){6-7}
 &  &  & \textbf{Accuracy} & \textbf{Loss} & \textbf{Accuracy} & \textbf{Loss} &  \\
\midrule
{BraTS19}
&    0.01          & D/F           & 0.915         & 0.232          & 0.921          & 0.181          & 110                         \\
&    0.001         & D/F           & 0.955         & 0.120          & 0.969          & 0.131          & 119                         \\
&    0.0001        & D\&F          & 0.982         & 0.110          & 0.970          & 0.120          & 125                         \\
{BraTS2020}
&   0.01           & D/F           & 0.939         & 0.200          & 0.939          & 0.171          & 108                         \\
&   0.001          & D/F           & 0.960         & 0.175          & 0.966          & 0.130          & 121                         \\
&   0.0001         & D\&F          & 0.981         & 0.121          & 0.974          & 0.112          & 127                         \\
{BraTS2021}
&   0.01           & D/F           & 0.959         & 0.137          & 0.932          & 0.153          & 115                         \\
&   0.001          & D/F           & 0.973         & 0.128          & 0.969          & 0.120          & 121                         \\
&   0.0001         & D\&F          & 0.985         & 0.103          & 0.973          & 0.116          & 126                         \\
\midrule
\end{tabular}
\vspace{-10pt}
\end{table*}

\subsection{Ablation Studies}
\label{subsec_ablation_studies}

\subsubsection{Model Architecture}
To evaluate the contribution of each architectural component in ReFRM3D, we conducted a component-wise ablation study on BraTS2021 using 100 randomly selected samples. These samples were randomly drawn from both HGG and LGG cases. {All experiments ran for 50 epochs} using the Adam optimizer at a learning rate of 0.0001, a batch size of 2, and a combined Dice and focal loss function.

Table \ref{tab:component_ablation} reports the detailed ablation experiments for four groups: Fusion: Fused Multi-scale Feature Fusion; Upsample: Hybrid Upsampling and Residual Integration (HURI); Skip: Residual Skip; and RadCond: Radiomic Conditioning. For the Fusion and the Upsample groups, the two components not under test are fixed to their best-performing variant. For the Skip group, fusion and upsampling are fixed to F5 and U5. For the Radiomic group, the backbone is fixed to F5, U5, and S5, so R2 through R4 share the same backbone and differ only in whether coarse-to-fine supervision and radiomic conditioning are added. Here, Trans+Res+Refine refers to transposed convolution, residual integration of encoder features, and a final refinement convolution.

\begin{table}[!ht]
\centering
\small
\caption{{Controlled optimizer comparison on BraTS2021 (train) with fixed input size and learning rate.}}
\label{tab:optimizer_comparison}
\begin{tabular}{llcccc}
\midrule
\multirow{2}{*}{\textbf{Metric}} & \multirow{2}{*}{\textbf{LR}} 
& \multicolumn{2}{c}{\textbf{$128\times128\times128$}} 
& \multicolumn{2}{c}{\textbf{$256\times256\times256$}} \\
\cmidrule(lr){3-4} \cmidrule(lr){5-6}
& & \textbf{Adam} & \textbf{SGD} & \textbf{Adam} & \textbf{SGD} \\
\midrule
Acc & 0.01   & 0.959 & 0.886 & 0.981 & 0.917 \\
& 0.001  & 0.973 & 0.929 & 0.984 & 0.960 \\
& 0.0001 & 0.985 & 0.941 & 0.990 & 0.967 \\
Loss & 0.01   & 0.137 & 0.156 & 0.162 & 0.181 \\
& 0.001  & 0.128 & 0.143 & 0.136 & 0.167 \\
& 0.0001 & 0.103 & 0.132 & 0.128 & 0.159 \\
T/E (s) & -- & 115--126 & 109--122 & 124--131 & 129--140 \\
\midrule
\end{tabular}
\end{table}

\begin{table*}[!ht]
\centering
\caption{Preprocessing component ablation. GPU memory and inference time are reported for a single volume forward pass. Training time (per epoch) and inference time are reported in the \textbf{t(s)/Ep} and \textbf{Inf/T(ms)} columns, respectively. Baselines are indicated with "$*$".}
\small
\label{tab:preproc_ablation}
\begin{tabular}{llcccccc}
\midrule
\textbf{Component} & \textbf{Variant} & \textbf{Tr/Acc} & \textbf{Tr/Loss} & \textbf{Avg DSC (\%)} & \textbf{t(s)/Ep} & \textbf{GPU Mem} & \textbf{Inf/T(ms)} \\
\midrule
{Slice Selection}
& Full range$^*$ & 0.891 & 0.27 & 81.34 & 151±2.1 & 2,847 MB & 312±8 \\
& Tumor-only  & 0.929 & 0.21 & 83.89 & 123±2.4 & 2,413 MB & 264±5 \\
{Normalization}
& No normalization$^*$         & 0.891 & 0.27 & 81.34 & 151±2.1 & 2,847 MB & 312±8 \\
& Min-max                   & 0.911 & 0.23 & 82.54 & 151±1.9 & 2,847 MB & 312±7 \\
& Z-score         & 0.929 & 0.20 & 83.12 & 151±1.7 & 2,847 MB & 312±6 \\
{Brain Isolation}
& No isolation$^*$            & 0.891 & 0.27 & 81.34 & 151±2.1 & 2,847 MB & 312±8 \\
& Axis-aligned              & 0.931 & 0.28 & 84.21 & 141±1.8 & 2,631 MB & 289±6 \\
& PCA (1.50$\sigma$) & 0.944 & 0.17 & 85.73 & 123±2.4 & 2,413 MB & 264±5 \\
\midrule
\end{tabular}
\vspace{-10pt}
\end{table*}

\begin{table}[!ht]
\centering
\small
\caption{{Sensitivity analysis of brain region cropping strategy and margin parameter on segmentation performance and training efficiency. Training time (per epoch) is reported in the \textbf{t(s)/Ep} column.}}
\label{tab:cropping_sensitivity}
\begin{tabular}{lcccc}
\midrule
\textbf{Cropping} & \textbf{Acc} & \textbf{Loss} & \textbf{DSC (\%)} & \textbf{t(s)/Ep} \\
\midrule
Axis-aligned        & 0.931 & 0.28 & 84.21 & 141 \\
PCA (0.75$\sigma$)  & 0.938 & 0.23 & 84.89 & 108 \\
PCA (1.00$\sigma$)  & 0.943 & 0.19 & 85.34 & 112 \\
PCA (1.25$\sigma$)  & 0.944 & 0.17 & 85.73 & 119 \\
\rowcolor{gray!20}
PCA (1.50$\sigma$)  & 0.944 & 0.17 & 85.73 & 123 \\
PCA (1.75$\sigma$)  & 0.941 & 0.21 & 80.89 & 129 \\
PCA (2.00$\sigma$)  & 0.936 & 0.23 & 77.12 & 135 \\
PCA (2.25$\sigma$)  & 0.933 & 0.27 & 74.31 & 138 \\
\midrule
\end{tabular}
\vspace{-5pt}
\end{table}

\subsubsection{Hyperparameter Configurations}
\label{hyperparams_x}
In this analysis of the experiments, we thoroughly examined how different configurations (i.e., image sizes, learning rates, optimizers, and loss functions) impacted the training and validation processes. 

The experiments were conducted over 100 epochs with a batch size of 2 for all configurations. Starting with image size, models trained with a smaller image size of $128^3$ generally resulted in faster convergence, as evidenced by lower training and validation loss and higher accuracy, compared to those trained with a larger image size of $256^3$ (Table \ref{tab:hyperparameter_configurations} reports different configurations). 

The choice of optimizer also had a significant impact. The Adam optimizer consistently outperformed SGD, with Adam maintaining lower training losses across all learning rates and input resolutions. We observed that the combination of Dice Loss and Focal Loss (D\&F) provided better stability across all datasets compared to using either Dice Loss or Focal Loss alone, i.e., the element-wise maximum of the two losses (denoted as D/F). We also conducted a controlled comparison on BraTS2021 to isolate the effect of the optimizer (see Table \ref{tab:optimizer_comparison}).




The impact of configurations also extended to training time, where data with $256^3$ image dimensions required more time per epoch (compared to $128^3$), regardless of the selected optimizer.

\vspace{-5pt}
\subsubsection{Preprocessing Configurations}
\label{preproc_slice}
Following the hyperparameter settings (see Section \ref{hyperparams_x}), we conducted a component-wise breakdown of the preprocessing pipeline, followed by a sensitivity analysis on cropping strategy and margin, using a subset of 70 random samples from BraTS2021.

Table \ref{tab:preproc_ablation} shows two distinct patterns. Brain isolation and slice selection change the input volume size. Consequently, their variants differ across all computational metrics. Normalization operates on the already cropped input. Thus, its variants share the same computational cost. Against the shared baseline, PCA-based isolation has the largest DSC gain and the largest reduction in training time; axis-aligned cropping recovers only part of this gain. Z-score normalization also achieves higher DSC compared to min-max normalization. Slice selection also provides a computational benefit. Processing only tumor-containing slices, roughly 68 on average instead of all 155 slices per volume, reduces computational overhead by approximately 56\% (see Figure \ref{fig:slice_range}).

Table \ref{tab:cropping_sensitivity} reports the sensitivity of the cropping and margin parameters. Here, the axis-aligned baseline scores the lowest accuracy and the highest training time per epoch, since its fixed orientation (in axial, coronal, and sagittal views) retains a larger non-brain region. 

Among the PCA-based configurations, accuracy and DSC improve steadily up to 1.25$\sigma$ and then stabilize through 1.50$\sigma$, where both reach their peak. Training time increases steadily across the range, as a larger margin produces a larger cropped volume. Beyond 1.50$\sigma$, performance degrades as the margin begins to reintroduce non-brain regions. We selected 1.50$\sigma$ as our final choice, since it achieves the same peak performance as 1.25$\sigma$ with a more conservative margin for anatomically variable cases.

\section{Discussion}
\label{sec_discussion}
The central idea of this work is that radiomic descriptors can guide segmentation directly. Prior radiomics-guided approaches extract features after segmentation is complete, which limits their use in a deployable pipeline. We instead extract descriptors from an intermediate coarse mask that the network produces internally, then use them to condition the final prediction.

The backbone was built with this goal, since an efficient backbone is a prerequisite for the radiomics pathway to be beneficial. FMFF fuses features across encoder depths so the model retains both fine boundary detail and broader tumor context. HURI pairs transposed convolution with a residual connection and recovers detail that plain interpolation loses. rSkip applies a learnable filter to encoder features before they reach the decoder. These choices together keep the model efficient at 8.64M parameters.

Preprocessing plays a similar supporting role. Brain isolation through PCA-based cropping adapts to the actual orientation of the brain rather than assuming a fixed axis. Additionally, the slice selection restricts training to tumor-containing regions only, which reduces computational overhead (see Section \ref{preproc_slice}) without compromising segmentation quality.

The ablation study isolates what radiomic conditioning provides. Coarse-to-fine supervision alone raises mean Dice from 84.13\% to 87\%. Radiomic conditioning on top of this raises it further to 88.13\%. This suggests that texture and intensity descriptors carry information that the convolutional features do not fully capture. External evaluation results on the BraTS-Africa dataset also indicate the generalizability of the model. Even with differences in scanner field strength, acquisition conditions, and patient demographics, the model reached an average DSC of 87.8\%.

Despite the effectiveness of the proposed model, a few limitations remain. For instance, our work relies on a supervised learning pipeline, which requires voxel-wise annotated data. Our current study is also restricted to adult pre-operative glioma cases. Similarly, pediatric brain tumor profiles differ considerably in morphology and radiomics characteristics, and would require targeted adaptation before deployment.

For future work, we aim to develop a more efficient framework that extends to post-operative and pediatric cases, which will involve incorporating a domain learning approach for unsupervised adaptation of the segmentation tasks. We will also explore the possibility of its integration into clinical workflows.

\section{Conclusion}
\label{conclusion}
In this paper, we developed a 3D MRI-based model for glioma segmentation. The model extends the 3D U-Net architecture with a fused residual backbone that incorporates Fused Multi-scale Feature Fusion, Hybrid Upsampling and Residual Integration, and Residual Skip Connections. We also introduced a novel radiomic conditioning mechanism that extracts texture and intensity descriptors from an intermediate coarse segmentation and re-injects them into the decoder to guide the final prediction. Experiments on the BraTS benchmark datasets (2019--2021) showed that the proposed model consistently achieved DSC values above 90\%. We further evaluated the model on the BraTS-Africa dataset to examine its performance across different imaging settings. The model maintained strong segmentation performance despite differences in scanner field strength and acquisition protocols. Our preprocessing methods also reduced resource consumption without substantially affecting segmentation performance. These findings suggest that ReFRM3D may serve as a useful research framework for automated glioma segmentation and subsequent quantitative tumor analysis.

\section*{Declarations}
\noindent
\textbf{Conflict of interest/Competing interests:} The authors declare no financial conflicts of interest or personal relationships that could have influenced this work and have consented to its publication.\\
\textbf{Data Availability:} The study uses four datasets: BraTS2019, BraTS2020, BraTS2021, and BraTS-Africa, all of which are publicly available.

\section*{References}

\end{document}